\PassOptionsToPackage{x11names,dvipsnames,table}{xcolor}

\documentclass[11pt, a4paper, google]{google}

\graphicspath{{./}{figure/}}

\usepackage[authoryear, sort&compress, round]{natbib}
\def\eqref#1{equation~\ref{#1}}

\def\1{\bm{1}}

\DeclareMathAlphabet{\mathsfit}{\encodingdefault}{\sfdefault}{m}{sl}
\SetMathAlphabet{\mathsfit}{bold}{\encodingdefault}{\sfdefault}{bx}{n}

\usepackage{bm}
\usepackage{mathtools}
\usepackage{mathrsfs}
\usepackage{dsfont}
\usepackage{nicefrac}
\usepackage{latexsym}

\usepackage{array}
\usepackage{multirow}
\usepackage{makecell}
\usepackage{arydshln}          
\usepackage{multicol}
\usepackage{paracol}
\usepackage{wrapfig}
\usepackage{subcaption}
\usepackage{threeparttable}

\usepackage[most]{tcolorbox}
\usepackage{stackengine}
\usepackage{tikz}
\usepackage{pgffor}
\usepackage{listings}
\usepackage{algorithm}
\usepackage{algpseudocode}

\usepackage{setspace}
\usepackage{ragged2e}
\usepackage[normalem]{ulem}
\usepackage{xspace}
\usepackage{lineno}
\usepackage[textsize=tiny]{todonotes}

\usepackage[capitalize,noabbrev]{cleveref}

\setlist[itemize]{leftmargin=*,noitemsep}

\tcbuselibrary{theorems}
\tcbuselibrary{breakable}
\newtcbtheorem[number within=section]{mytheo}{Definition}%
{colback=red!5,colframe=red!35!black,fonttitle=\bfseries}{th}

\definecolor{added}{rgb}{0,0.6,0}
\definecolor{deleted}{rgb}{1,0,0}

\newcommand{\hmap}[2]{%
    \pgfmathsetmacro{\percent}{((#1 - 1) / 4) * 100}%
    \ifdim \percent pt > 50 pt
        \def\mytextcol{white}%
    \else
        \def\mytextcol{black}%
    \fi
    \edef\temp{\noexpand\cellcolor{green!\percent!white}}\temp%
    \textcolor{\mytextcol}{$#1_{#2}$}%
}

\theoremstyle{plain}

\theoremstyle{definition}

\theoremstyle{remark}

\definecolor{verylightgray}{gray}{0.9}

\definecolor{light_red}{rgb}{1.0, 0.6, 0.6}
\definecolor{light_green}{rgb}{0.56, 0.93, 0.56}

\newcommand{\xhdr}[1]{\noindent\textbf{#1.}}

\newcommand{\best}{\cellcolor{green!25}\bfseries}
\newcommand{\second}{\cellcolor{blue!10}}

\newcommand{\LLMbare}{LLM (No Events)}
\newcommand{\LLMev}{LLM (Event RAG)}
\newcommand{\SEERnoK}{SEER w/o Knowledge}
\newcommand{\SEERfull}{SEER}

\uselogo{}

\title{SEER: Self-Evolving Event Reasoning \\ and Retrieval for Time Series Forecasting}


\correspondingauthor{wtd3gz@virginia.edu, mihirparmar@google.com, palashgoyal@google.com}

\renewcommand{\today}{}

\author[1,2]{\fontsize{10.0pt}{10.0pt}\selectfont Mingtian Tan}
\author[1]{\fontsize{10.0pt}{10.0pt}\selectfont Palash Goyal}
\author[1]{\fontsize{10.0pt}{10.0pt}\selectfont Mihir Parmar}
\author[1]{\fontsize{10.0pt}{10.0pt}\selectfont Sarkar Snigdha Sarathi Das}
\author[1]{\fontsize{10.0pt}{10.0pt}\selectfont Chun-Liang Li}
\author[1]{\fontsize{10.0pt}{10.0pt}\selectfont Nanyun Peng}
\author[2]{\fontsize{10.0pt}{10.0pt}\selectfont Thomas Hartvigsen}
\author[1]{\fontsize{10.0pt}{10.0pt}\selectfont Jinsung Yoon}
\author[1]{\fontsize{10.0pt}{10.0pt}\selectfont Tomas Pfister}

\newsavebox{\googlelogobox}
\sbox{\googlelogobox}{%
  \includegraphics[height=1.5ex]{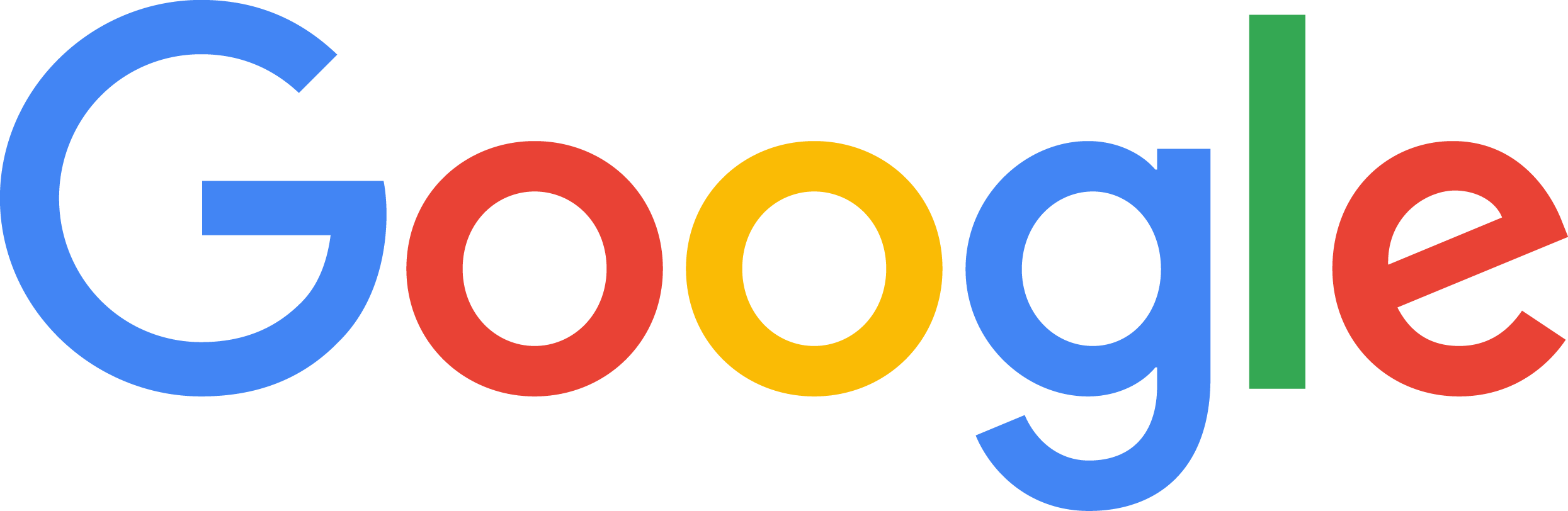}%
}
\DeclareRobustCommand{\googlelogo}{%
  \raisebox{-0.3ex}{\usebox{\googlelogobox}}%
}

\newsavebox{\uvalogobox}
\sbox{\uvalogobox}{%
  \includegraphics[height=1.7ex]{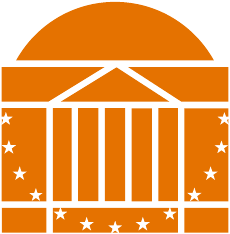}%
}
\DeclareRobustCommand{\uvalogo}{%
  \raisebox{-0.3ex}{\usebox{\uvalogobox}}%
}

\affil[1]{\fontsize{9.0pt}{9.0pt}\selectfont \protect\googlelogo\hspace{0.1em} Google Cloud AI Research}
\affil[2]{\fontsize{9.0pt}{9.0pt}\selectfont \protect\uvalogo\hspace{0.1em} University of Virginia}

\begin{abstract}
Real-world time series are frequently driven by exogenous events and structural shifts, rendering conventional forecasting based solely on historical numerical observations insufficient. While language models can retrieve external news, standard retrieval-augmented approaches struggle with high noise, missing signals, and an inability to reason causally about event impacts. We propose SEER (Self-Evolving Event Reasoning and Retrieval), a closed-loop framework that dynamically optimizes event conditioning for time series forecasting. SEER translates prediction errors into two decoupled feedback mechanisms: (i) a reflective retrieval memory that refines subsequent search queries and filters spurious noise, and (ii) a persistent causal knowledge base that distills transferable domain dynamics. SEER enforces strict chronological boundaries across both event retrieval and reflection, preventing look-ahead bias and data leakage. Across six volatile time-series benchmarks, SEER consistently outperforms state-of-the-art time series foundation models and language model baselines.

\vspace{4pt}
{\centering
\raisebox{-0.16\height}{\includegraphics[height=1.05em]{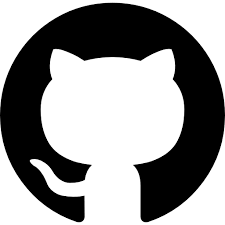}}\hspace{0.32em}\href{https://github.com/BennyTMT/SEER/}{\texttt{https://github.com/BennyTMT/SEER}}%
\par}

\end{abstract}

\begin{document}

\maketitle

\begin{figure}[H]
    \centering
    \includegraphics[width=.98\linewidth]{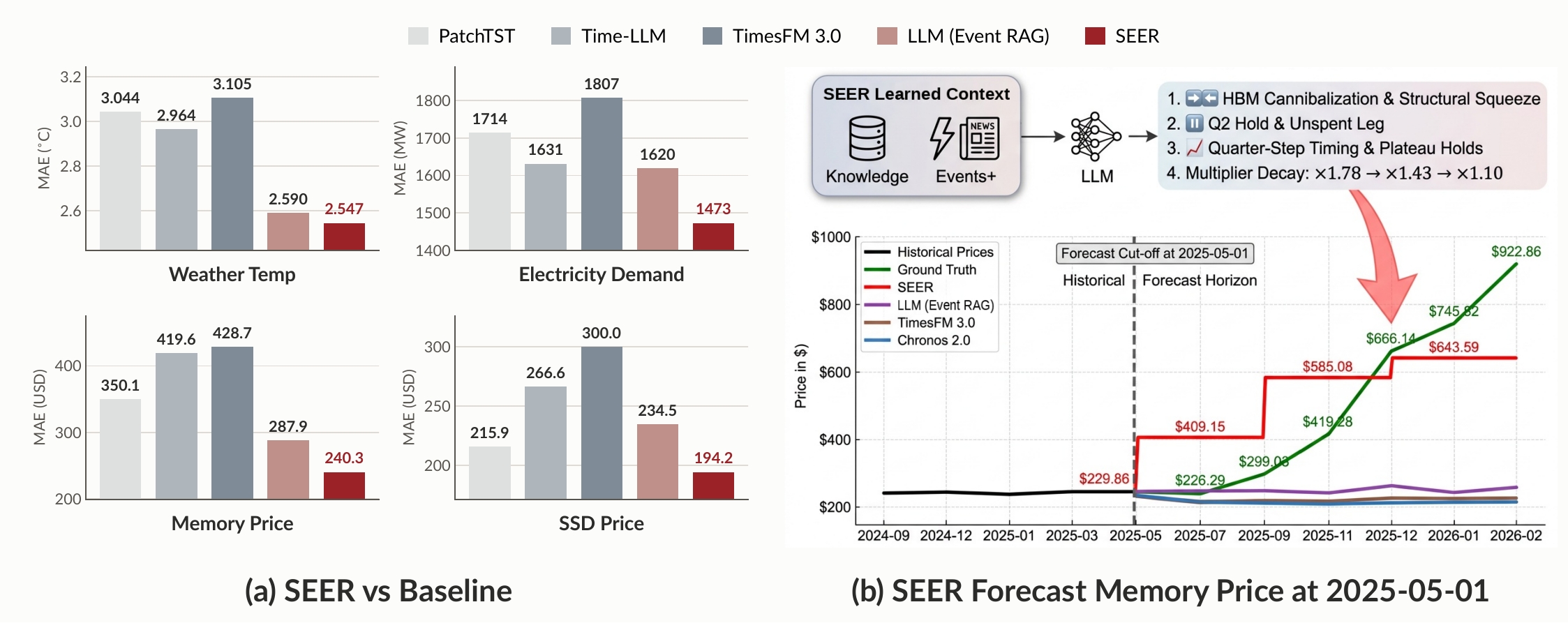}
    \caption{\textbf{Performance and interpretability of SEER.} {(a)} SEER demonstrates consistent performance improvements over prior forecasting paradigms across event-driven time series tasks (e.g., Memory/SSD prices, Electricity, and Weather). Detailed results and comparisons with additional baselines are provided in ~\autoref{tab:main-tsf} (\textsection\ref{sec:exp-tsf}). {(b)} A memory price forecasting case study (\emph{64GB-DDR5} from \citet{pangolyPriceTrends}) illustrates SEER's interpretable predictions, driven by dynamic event retrieval and causal knowledge discovery. For details, see \autoref{fig:64gb-ddr5-prompt} and \ref{fig:64gb-ddr5-reasoning} in Appendix \ref{app:prompts-cases}.}
    \label{fig:seer_demo}
\end{figure}

\section{Introduction}
\label{sec:introduction}

Time series forecasting drives critical decision-making across finance~\citep{xiao2025trade,fan2025ai}, weather~\cite{xu2025fidel,williams2024context}, energy management~\citep{wang2024newsforecast,ferdaus2025foundation}, and supply chain logistics~\citep{cristian2024interseries,schmid2025comparing}.
Deep learning architectures~\cite{Yuqietal-2023-PatchTST,liu2023itransformer,wu2021autoformer} and massively pre-trained Time Series Foundation Models (TSFMs)~\cite{das2024decoder,ansari2024chronos} have achieved remarkable success by learning complex temporal dependencies, periodicities, and local patterns from historical observations.
However, real-world systems are rarely closed.
They are continually subject to exogenous shocks such as geopolitical shifts, policy interventions, corporate earnings surprises, and supply disruptions.
In these non-stationary regimes, historical numerical observations alone are fundamentally insufficient; the signals governing future trajectories reside outside the numerical sequence.

To capture these exogenous drivers, recent efforts leverage language models to integrate unstructured textual data into the forecasting pipeline~\cite{jin2023time, zhou2024one,xue2023promptcast}.
Most existing paradigms~\citep{tan2024language,yang2026prophet} adopt static retrieval-augmented generation (RAG), querying external news corpora or financial feeds to supplement historical values with contemporaneous text.
While intuitive, standard RAG architectures face three fundamental failure modes when applied to time-series forecasting: i) \textit{low signal-to-noise ratio:} raw news streams are saturated with tangential, redundant, or misleading reports, ii) \textit{absence of causal reasoning:} standard RAG relies on semantic proximity rather than causal attribution, and iii) \textit{open-loop inefficiency:} static approaches do not leverage forecast errors to adjust context.

To bridge this gap, we present \textbf{Self-Evolving Event Reasoning and Retrieval (SEER)}, a closed-loop framework that dynamically optimizes event conditioning for time-series forecasting.
Rather than treating event integration as a one-shot retrieval task, SEER continuously audits forecast discrepancies against realized trajectories, converting residual prediction errors into two decoupled feedback mechanisms: i) \textit{reflective retrieval memory:} an episodic memory bank that evaluates whether past retrieval queries succeeded in capturing the true drivers behind an observed movement, and ii) \textit{persistent causal knowledge base:} a semantic repository that abstracts episodic forecasting failures into persistent, transferable causal mechanisms.
Crucially, all knowledge base updates and reflective memories are indexed strictly up to time step $t$, eliminating the temporal leakage and look-ahead bias.
Across six volatile, event-driven tasks (including components prices, electricity, weather, stocks and prediction markets), we show that SEER significantly outperforms prior forecasting paradigms while exhibiting interpretable predictive behaviors (see \autoref{fig:seer_demo}). Furthermore, our ablations reveal that (1) SEER's self-evolving distilled knowledge effectively reduces forecasting errors, and (2) compared to standard raw event retrieval, SEER's event refinement consistently improves performance across multiple backbone LLMs. In this work, we make the following contributions:
\begin{itemize}
    \item We propose a novel closed-loop event conditioning paradigm for time series forecasting. Instead of static RAG, we formulate event conditioning as an error-driven, closed-loop process where predictive residuals continuously refine external knowledge acquisition (\textsection\ref{sec:method-env}).
    
    \item We design the SEER framework with two complementary feedback mechanisms - a reflective retrieval memory for dynamic query refinement and noise suppression, and a persistent causal knowledge base for abstracting transferable structural dynamics (\textsection\ref{sec:method-learn}).

    \item We perform comprehensive experiments to show that across six diverse and volatile time-series benchmarks, SEER consistently outperforms state-of-the-art numerical time-series foundation models, deep sequence architectures, and language model baselines (\textsection\ref{sec:exp-tsf} and \ref{sec:exp-non-tsf}).
\end{itemize}

\section{Related Work}

\xhdr{Time Series Forecasting Methods} Recent advancements in time series forecasting (TSF) mainly follow three distinct trajectories. {1) Deep Learning-based Numerical Forecasters:} Methods such as Autoformer~\cite{wu2021autoformer}, DLinear~\cite{zeng2023transformers}, and PatchTST~\cite{Yuqietal-2023-PatchTST} are supervised architectures trained on specific datasets. By explicitly capturing domain-level temporal characteristics, they consistently establish strong performance baselines within their target domains. {2) Time Series Foundation Models (TSFMs):} Pre-trained on massive, multi-domain datasets encompassing weather, electricity, finance, and synthetic data, models such as TimesFM 3.0~\cite{das2024decoder} and Chronos-2~\cite{ansari2024chronos} provide robust zero-shot forecasting capabilities without task-specific fine-tuning. Both DL-based forecasters and TSFMs share a fundamental constraint: they rely exclusively on numerical inputs. To address this, {3) LLM-Based Forecasters:} Methods including PromptCast~\cite{xue2023promptcast}, One-Fits-All~\cite{zhou2024one}, and Time-LLM~\cite{jin2023time} integrate pre-trained LLMs (e.g., GPT-2) as the primary forecasting backbone for either inference or fine-tuning. These approaches typically append static textual information, such as domain descriptions and statistical metadata, to assist prediction during inference. However, \citet{tan2024language} demonstrate that the effectiveness of these methods is limited, due to both the absence of meaningful exogenous signals, such as the actual events driving time series variations, and the failure of static domain information to provide actionable predictive insights for the LLMs.

\xhdr{Self-Evolving Agents and Dynamic Environments} Recent advancements in Large Language Model (LLM) agents have focused on mechanisms for self-reflection and experience accumulation to minimize human supervision~\cite{fang2025comprehensive, gao2025survey}. Works such as Reflexion~\cite{shinn2023reflexion} introduced the concept of verbal reinforcement, where agents reflect on past trajectories to improve future reasoning. To support long-horizon interaction, various memory selection mechanisms~\cite{packer2023memgpt, zhong2024memorybank} and consolidation strategies~\cite{wang2025agentworkflowmemory, nottingham2024skillset} have been introduced. However, moving beyond the mere archiving of raw successful traces, contemporary research increasingly emphasizes generalized experiential learning. More recently, frameworks like ReasoningBank~\cite{ouyang2026reasoningbank} and other memory-augmented agents~\cite{zhao2024expel, zheng2024synapse} distill structured actionable knowledge into reflective memories to avoid repeating past mistakes. By actively extracting transferable strategies from historical success or failure~\cite{zhang2024context, alazraki2025no}, these systems generalize to unseen tasks rather than rigidly executing memorized routines. 

Concurrently, there is a growing interest in moving beyond rigid, static testing environments, which are increasingly bottlenecked by benchmark overfitting~\cite{zhang2025swebenchgoeslive, hu2026seal}. To mitigate this, recent studies~\cite{zala2024envgen, pan2024training} leverage code-driven synthesis and LLM-based world models~\cite{zuo2026qwen, wang2026agentworldmodel} to construct scalable, interactive settings. Advancing this paradigm, frameworks like EnvHarness~\cite{huang2026envharness} and GenEnv~\cite{guo2025genenv} introduce adaptive environment generation, automatically mutating tasks to synthesize more diverse and challenging environments for training robust agents. In contrast, SEER employs an evolving paradigm to dynamically enrich the forecasting environment, mitigating the issue that existing information is often insufficient to support accurate predictions.

\section{Method}
\label{sec:method}



\subsection{Problem Setup}
\label{sec:method-setup}
We frame the forecasting task sequentially over discrete time cut-offs $t=1,\dots,T$ for a given set of entities (e.g., stock prices, electricity demand, or prediction markets). At each step $t$, the base query $x_t$ comprises the entity description, the target forecasting horizon $H$, and the historical time-series data within a specific look-back window $[t-L, t]$. The objective is to predict the corresponding outcome $y_t$ over the subsequent period $(t, t+H]$. To enable sequential learning while strictly preventing look-ahead bias, we chronologically partition the sequence into a training set ($t \le \tau$) and a testing set ($t > \tau$). During the learning phase, SEER iteratively optimizes the task loss $\ell$ using only past trajectories whose outcomes are merely observed within the training set (i.e., $t+H \le \tau$).

\begin{figure}[H]
    \centering
    \includegraphics[width=.98\linewidth]{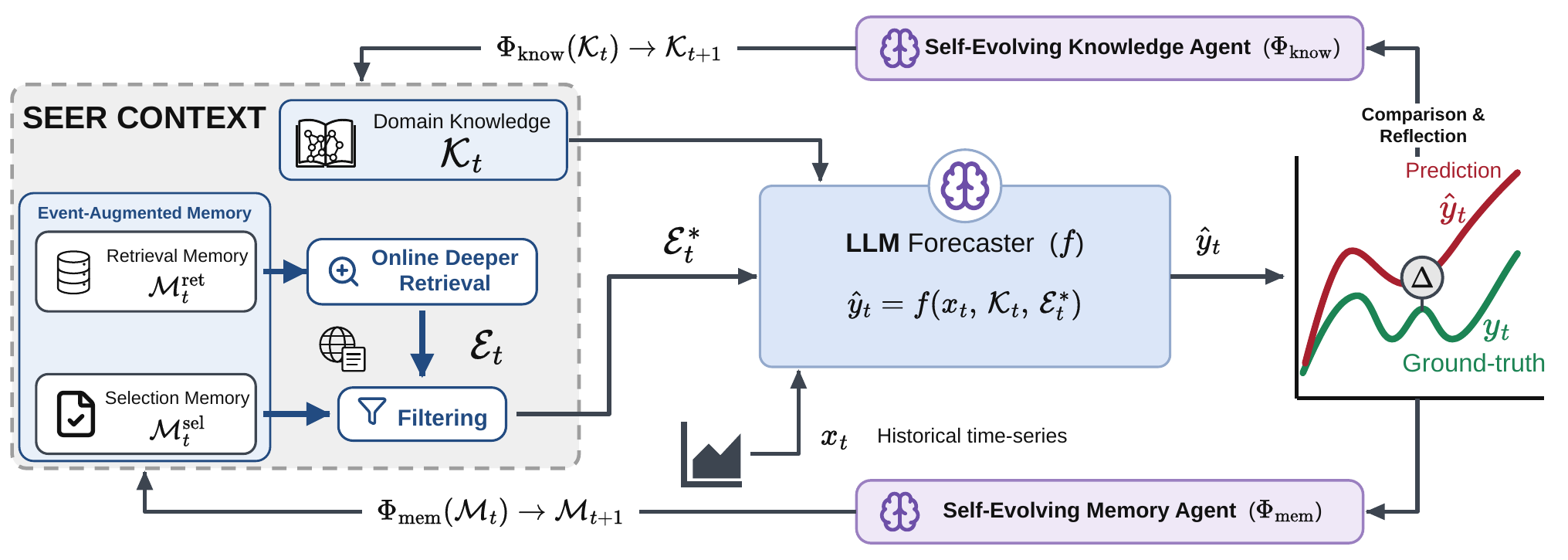}
    \caption{Overview of the SEER framework. At step $t$, an LLM Forecaster ($f$) generates a reasoning process $r_t$, which contains the prediction $\hat{y}_t$, based on historical numerical data $x_t$ and a dynamic textual context space. This context space supplies the LLM with augmented external events (further retrieved using $\mathcal{M}_t^{\text{ret}}$ and filtered via reflective memory $\mathcal{M}_t^{\text{sel}}$) alongside persistent causal knowledge $\mathcal{K}_t$. Upon observing the true outcome $y_t$, SEER closes the feedback loop by reflecting on the prediction error. The updates are decoupled: $\Phi_{\mathrm{mem}}$ refines the Event-Augmented Memory $\mathcal{M}_{t+1}$, while $\Phi_{\mathrm{know}}$ extracts transferable domain dynamics from past errors to form $\mathcal{K}_{t+1}$.}
    \label{fig:seer_pipeline}
\end{figure}

A frozen backbone language model (LM), denoted as $f$, maps the query $x_t$ and a textual context to a prediction. Because $f$ remains frozen during inference, the core challenge shifts from updating model parameters to optimizing the context provided to the model. SEER addresses this by learning to dynamically retrieve and refine the most predictive exogenous information.

\subsection{Forecasting with an Evolving Textual Context}
\label{sec:method-env}

A query relying solely on the endogenous information $x_t$ is often insufficient, as it fails to capture the exogenous events driving real-world dynamics. We therefore formulate forecasting as an agent reasoning over a structured textual context:

\begin{equation}
\Omega_t \;=\; \bigl(\mathcal{E}_t,\ \mathcal{K}_t\bigr),
\label{eq:env}
\end{equation}

where $\mathcal{E}_t$ represents the local pool of external events available at cut-off $t$, and $\mathcal{K}_t$ is the domain causal forecasting knowledge distilled from past forecasting attempts. 

Unlike static knowledge bases, both components are represented in natural language and are dynamically updatable over time. In our framework, the context space $\Omega_t$ undergoes continuous closed-loop learning, enabling the core mechanism of \textbf{Self-Evolving Event Reasoning and Retrieval (SEER)}.

To process this context space, as shown in \autoref{fig:seer_pipeline}, the agent maintains a reflective retrieval memory $\mathcal{M}_t$ and a persistent causal knowledge base $\mathcal{K}_t$. Together, these components govern a context construction operator $R$, yielding the final forecast:
\begin{equation}
\hat{y}_t \;=\; f\bigl(x_t,\ R(\mathcal{E}_t,\mathcal{K}_t, \mathcal{M}_t)\bigr).
\label{eq:seer}
\end{equation}

The operator $R$ constructs the forecasting context by supplying the LM with both refined events and causal knowledge. To achieve this, the reflective memory $\mathcal{M}_t$ is explicitly decoupled into two components: retrieval memory $\mathcal{M}_t^{\text{ret}}$ and selection memory $\mathcal{M}_t^{\text{sel}}$. Specifically, $R$ utilizes $\mathcal{M}_t^{\text{ret}}$ to dynamically expand the relevant event pool, applies $\mathcal{M}_t^{\text{sel}}$ to filter out noise, and outputs the refined events alongside the causal forecasting knowledge $\mathcal{K}_t$. Formally:

\begin{equation}
R(\mathcal{E}_t,\mathcal{K}_t, \mathcal{M}_t)
\;=\;
\bigl(\underbrace{\rho_{\mathcal{M}_t^{\text{sel}}} \circ \sigma_{\mathcal{M}_t^{\text{ret}}}(\mathcal{E}_t)}_{\text{filtered events}},\;
\underbrace{\mathcal{K}_t}_{\text{causal knowledge}}\bigr).
\label{eq:read}
\end{equation}

These operations are driven by two continuously updated mechanisms:

\begin{itemize}
    \item \textbf{Memory-Guided Expansion and Filtering:} The expansion operator $\sigma_{\mathcal{M}_t^{\text{ret}}}$ utilizes the retrieval memory $\mathcal{M}_t^{\text{ret}}$ to actively formulate search queries and gather missing exogenous information, thereby dynamically expanding the event pool $\mathcal{E}_t$. Subsequently, its selection component $\mathcal{M}_t^{\text{sel}}$ guides the operator $\rho_{\mathcal{M}_t^{\text{sel}}}$ to filter the expanded event pool and remove noise.
    
    \item \textbf{Knowledge-Driven Causal Reasoning:} Rather than relying on static logic, the framework continuously distills and updates the causal forecasting knowledge $\mathcal{K}_t$ from past forecasting attempts. This persistent knowledge provides direct reasoning support to the backbone $f$.
\end{itemize}

\subsection{Self-Reflective Evolution}
\label{sec:method-learn}

During the learning phase, once the ground-truth outcome $y_t$ becomes observable, SEER closes the feedback loop via a verbal reflection mechanism. Specifically, the agent analyzes the execution trajectory $\xi_t = (x_t, R(\mathcal{E}_t, \mathcal{K}_t, \mathcal{M}_t), r_t, y_t)$, where $r_t$ denotes the intermediate reasoning steps and includes the final prediction $\hat{y}_t$. By comparing the reasoning process $r_t$ with the ground truth $y_t$, the agent diagnoses the causes of the prediction error $\ell(\hat{y}_t, y_t)$.

This diagnosis is directly distilled into textual updates for the context space. To prevent interference between retrieval deficiencies (inadequate evidence) and reasoning flaws (flawed logic), SEER translates the residual errors into two decoupled feedback mechanisms via distinct reflection operators:
\begin{align}
\mathcal{M}_{t+1}
  &\;=\; \mathcal{C}\circ\Phi_{\mathrm{mem}}\bigl(\xi_t,\ \mathcal{M}_t\bigr),
\label{eq:reflect-mem}\\[2pt]
\mathcal{K}_{t+1}
  &\;=\; \mathcal{C}\circ\Phi_{\mathrm{know}}\bigl(\xi_t,\ \mathcal{K}_t\bigr).
\label{eq:reflect-know}
\end{align}
Here, $\Phi_{\mathrm{mem}}$ updates $\mathcal{M}_t$ to refine future search queries ($\mathcal{M}_t^{\text{ret}}$) and noise filters ($\mathcal{M}_t^{\text{sel}}$), while $\Phi_{\mathrm{know}}$ derives general causal forecasting rules from past errors to update the persistent knowledge base $\mathcal{K}_t$. The detailed reflection prompt template are provided in~\autoref{fig:seer-learn-mm} and \ref{fig:seer-learn-know} in Appendix~\ref{app:prompts-cases}.

Both operators initially append new insights. To manage context length limits and prevent contextual redundancy, they are followed by a condensation operator $\mathcal{C}$. When $\mathcal{M}$ or $\mathcal{K}$ approaches its text capacity, $\mathcal{C}$ prompts the LLM to merge highly similar entries and compress redundant texts, thereby maintaining a concise and informative set of memories and causal rules. The prompt used for this condensation is provided in~\autoref{fig:seer-condense} of Appendix~\ref{app:prompts-cases}.

Through sequential forecast-observe-reflect iterations over the training window, this closed-loop mechanism optimizes the textual components comprising the forecasting context space:

\begin{equation}
\min_{\mathcal{M},\,\mathcal{K}}\ \ \mathbb{E}\Bigl[\textstyle\sum_{t} \ell\bigl(y_t,\ \hat{y}_t\bigr)\Bigr].
\end{equation}

\textbf{Backtesting Protocol:} During evaluation on the testing set ($t > \tau$), the memory modules $\mathcal{M}$ and $\mathcal{K}$ are frozen. The agent performs inference solely via Equation~\eqref{eq:seer}. The initial event pool $\mathcal{E}_t$ is constructed through two rounds of general targeted searches, which are subsequently augmented by guided searches driven by $\mathcal{M}^{\text{ret}}$ (the retrieval component of $\mathcal{M}$). The corresponding prompt templates are detailed in \autoref{fig:seer-search-init} and \ref{fig:seer-search-missing} in Appendix~\ref{app:prompts-cases}. Notably, all retrieved exogenous events are strictly constrained to have occurred on or before the forecasting cut-off $t$ by leveraging the \emph{fact-checking filtering mechanism} of LEAF~\cite{tan2026leaf}, whose reliability is verified through human expert audit. This chronological constraint explicitly prevents look-ahead bias.


\section{Experiments}
\label{sec:experiments}

We evaluate SEER's performance across six  domains (five time-series tasks and one event-forecasting task) to answer four research questions: \textbf{(RQ1)} Does SEER outperform state-of-the-art time-series forecasting (TSF) methods? \textbf{(RQ2)} Does it show a clear improvement over its backbone LLM (with or without basic event retrieval)? \textbf{(RQ3)} Can it generalize beyond TSF to broader forecasting tasks? \textbf{(RQ4)} How does the learned knowledge contribute to forecasting performance?

\subsection{Experimental Setup}
\label{sec:exp-intro}

\paragraph{Data and Evaluation.}\autoref{tab:data} summarizes the six evaluation domains, detailing their sampling frequencies, horizons, forecasting targets, and metrics. Each domain is strictly split into learning and evaluation windows to construct a backtesting environment. Given that numerical time-series forecasting benchmarks (e.g., GIFT-Eval~\cite{aksu2024giftevalbenchmarkgeneraltime}, or Monash Archive~\cite{godahewa2021monash}) often include weather (temperature) and electricity (demand) predictions, we incorporate these two domains. We update the data using the Meteostat API~\citep{meteostat2026python} and the Grid Status API~\citep{kanter2026gridstatus} to forecast the daily maximum and minimum values across major US cities for the next 7 days; Stock covers 50 US equities, forecasting the next-day closing price direction (Up, Neutral, or Down); For event forecasting, we adopt a subset of questions from ForecastBench~\cite{karger2024forecastbench}, focusing primarily on the ``Tech'', ``Politics'', and ``Economy'' domains. We predict the probability of the final outcome at six timestamps, spanning from the market's creation to the point where 10\% of its duration remains before resolution; Memory and Solid-State Drive (SSD) track monthly memory component prices sourced from price tracking websites~\cite{pangolyPriceTrends,pcpartpickerPriceTrends}, forecasting prices over a 6- or 9-month horizon; Further dataset details, including specific stock tickers and the selected cities, are provided in Appendix~\ref{app:targets}.



For the Stock domain, we apply a stride of 2 days (excluding non-trading days), yielding 25 cut-off dates and 1,250 test samples. Similarly, Weather and Electricity use a 7-day stride, resulting in 220 and 66 samples, respectively. For Memory and SSD, the stride is set to one month; accounting for missing prices of components during the evaluation window, these yield 246 and 164 samples. Finally, Polymarket evaluates six forecasting cut-offs per question, totaling 276 test samples.

\begin{table*}[t]
\centering
\caption{\textbf{Summary of the six evaluation domains.} The table outlines the number of entities, chronological splits for the learning and evaluation windows, forecasting horizons, and target evaluation metrics for each task. Multiple horizons (e.g., 6 / 9 months) represent distinct experimental settings.}
\label{tab:data}
\resizebox{\textwidth}{!}{%
\begin{threeparttable}
\begin{tabular}{@{} l c l l c l @{}}
\toprule
\textbf{Domain} & \textbf{Entities} & \textbf{Learning Window} & \textbf{Evaluation Window} & \textbf{Horizon} & \textbf{Target (Metric)} \\
\midrule
Weather     & 10 & 3 seasonal ranges\tnote{\dag}  & 2026-02-22 -- 2026-07-19 & 7 days & Daily Temp (MAE, MSE) \\
Electricity & 3  & 3 seasonal ranges\tnote{\dag}  & 2026-02-22 -- 2026-07-19 & 7 days & Daily Demand (MAE, MSE) \\
Stock       & 50 & 2025-10-01 -- 2026-02-01       & 2026-02-01 -- 2026-05-01 & 1 day  & Daily Price Direction (Accuracy) \\
Polymarket  & 146 & Pre-split resol.\tnote{\ddag}  & Post-split resol.\tnote{\ddag} & To resol. & Probability (Brier Score) \\
Memory         & 20 & 2024-02-01 -- 2025-05-01       & 2025-05-01 -- 2026-06-01 & 6\,m / 9\,m & Monthly Price (MAE, MSE) \\
SSD         & 10 & 2024-02-01 -- 2025-05-01       & 2025-05-01 -- 2026-08-01 & 6\,m / 9\,m & Monthly Price (MAE, MSE) \\
\bottomrule
\end{tabular}
\begin{tablenotes}
  \footnotesize
  \item[\dag] Seasonal ranges: 2023-02-26 to 2023-08-04, 2024-02-26 to 2024-08-04, and 2025-02-24 to 2025-08-03.
  \item[\ddag] Split by resolution date: 2026-04-01 for Politics/Economy and 2026-01-01 for Tech, yielding 100 training events and 46 testing events.
\end{tablenotes}
\end{threeparttable}%
}
\end{table*}

\paragraph{Implementation.}
We instantiate SEER using four LLMs from~\citet{deepmind2026gemini31pro} and \citet{anthropic2025claudesonnet45}, including Gemini 3.1 Pro, Gemini 3.1 Flash, Claude 4.6 Sonnet, and Claude 4.8 Opus. As detailed in Section~\ref{sec:method}, we employ two rounds of general targeted searches to construct the initial event pool, which is subsequently augmented during the evaluation phase by an additional round of guided searches leveraging SEER's learned retrieval memory $\mathcal{M}^{\text{ret}}$. To ensure temporal authenticity and strictly prevent future information leakage, all retrieved events are processed through the \emph{fact-checking filtering mechanism} of LEAF~\cite{tan2026leaf}. The input data context window for forecasting is set to 14 days for Weather, Electricity, and Stock, and 9 months for component prices.

For the implementation of memory and knowledge, we adopt a shared context space. Within a given domain, this context is shared across different forecasting targets, leveraging the transferability of domain knowledge and target correlations. While the context space is shared, the learning process is sequential: the context is updated based on the forecasting feedback of a single sample before proceeding to the next. Additionally, due to the diversity of forecasting targets in certain domains, we partition the stock and event forecasting domains into 6 and 3 subdomains, respectively.

\paragraph{Selective Stock Forecasting}
Although predicting stock markets remains an open challenge~\cite{chen2025mtbench,wang2024newsforecast,dong2024fnspid_acm}, we assume that a small subset of price movements is still predictable. However, these limited predictable instances are often overwhelmed by vast unpredictable noise. Building on this assumption, we formulate a \emph{Selective Stock Forecasting} task. On any given cut-off date, the LLM is provided with the recent prices and retrieved events for all 50 equities. It is then tasked with selecting and forecasting at most six stocks ($\leq 6$ per day) that it deems predictable, retaining the option to abstain entirely. This setup requires the LLM to distinguish strong, actionable exogenous signals from pure noise or information already priced into the market.

\paragraph{Baselines.}
We compare SEER against three baseline families evaluated on identical testing sets. The trained baseline methods utilize the same training data as detailed in Table~\ref{tab:data}. \textbf{DL-based TSF} includes seven supervised models (Autoformer~\cite{wu2021autoformer}, DLinear~\cite{zeng2023transformers}, TimesNet~\cite{wu2023timesnet}, PatchTST~\cite{Yuqietal-2023-PatchTST}, Crossformer~\cite{zhang2023crossformer}, iTransformer~\cite{liu2023itransformer}, TimeMixer~\cite{wang2023timemixer}) trained purely on numeric series without language models. \textbf{LLM-based TSF} comprises six methods (PromptCast~\cite{xue2023promptcast}, One-Fits-All~\cite{zhou2024one}, Time-LLM~\cite{jin2023time}, CALF~\cite{liu2025calf}, TimeCMA~\cite{liu2025timecma}, TimeCAP~\cite{lee2025timecap}) adapting numeric series through LLMs (e.g., GPT-2 or Gemini-3.1-flash-lite). \textbf{Time-series foundation models (TSFMs)} include zero-shot forecasters TimesFM 3.0~\cite{das2024decoder} and Chronos-2~\cite{ansari2024chronos}, evaluated without task-specific fine-tuning. 

\subsection{Ablation Settings}
\label{sec:method-settings}
The decoupled design of our framework allows each component to be evaluated independently; to this end, we conduct our ablation study using the following configurations:
\begin{itemize}
    \item \LLMbare{}: Uses only the historical time-series data $x_t$, yielding $f(x_t)$.
    \item \LLMev{}: Incorporates the initial event pool $\mathcal{E}_t$ without memory-guided retrieval or causal knowledge, formulated as $f(x_t,\mathcal{E}_t)$.
    \item \SEERnoK{}: Employs memory-guided retrieval to augment and filter the generic pool while omitting causal knowledge, formulated as $f\bigl(x_t, \rho \circ \sigma(\mathcal{E}_t)\bigr)$.
    \item \SEERfull{}: The complete proposed framework, yielding $f\bigl(x_t, R(\mathcal{E}_t, \mathcal{K}_t, \mathcal{M}_t)\bigr)$.
\end{itemize}

\subsection{SEER's Performance on Time-Series Tasks}
\label{sec:exp-tsf}

\autoref{tab:main-tsf} compares SEER against 15 time-series forecasting (TSF) baselines and the backbone LLM (Claude 4.8 Opus) equipped with event RAG across five domains. SEER consistently outperforms all TSF categories, including supervised deep learning models, LLM-based forecasters, and zero-shot TSFMs. Notably, SEER achieves consistent error reductions, such as a 50.2\% MSE improvement over the best baseline on the Memory 6m task, and a 16.4\% reduction on Electricity. Furthermore, its relative improvements in MSE generally exceed those in MAE (e.g., a 50.2\% vs. 16.5\% reduction on Memory 6m), indicating SEER's effectiveness in mitigating extreme errors during sudden, event-driven shifts. Finally, SEER maintains a robust advantage at longer horizons (e.g., 9 months) where historical autoregressive signals typically decay, reducing the MSE by 24.3\% on Memory and 19.6\% on SSD compared to the best-performing traditional deep learning baselines.

\begin{table*}[t]
\centering
\renewcommand{\arraystretch}{1.1} 
\caption{Performance comparison between SEER and baselines on time-series forecasting tasks. MSE values for Memory, SSD, and Electricity are scaled by $10^{-3}$. For the Stock domain, we convert the exact price predictions of baseline methods into price directions to evaluate accuracy; in each column, \colorbox{green!15}{the best result} is shaded and set in bold, and \colorbox{blue!7}{the second best} is lightly shaded. Results including confidence intervals are detailed in \autoref{tab:main-tsf-ci} in Appendix~\ref{app:other-results}.}
\label{tab:main-tsf}
\small
\resizebox{\textwidth}{!}{%
\begin{tabular}{@{}l@{\hspace{4mm}} rr rr rr rr c rr rr @{}}
\toprule
\multirow{2}{*}{\textbf{Method}} 
& \multicolumn{2}{c}{\textbf{Memory 6M}} & \multicolumn{2}{c}{\textbf{Memory 9M}} 
& \multicolumn{2}{c}{\textbf{SSD 6M}} & \multicolumn{2}{c}{\textbf{SSD 9M}} 
& \textbf{Stock} 
& \multicolumn{2}{c}{\textbf{Weather}} & \multicolumn{2}{c}{\textbf{Electricity}} \\
\cmidrule(lr){2-3} \cmidrule(lr){4-5} \cmidrule(lr){6-7} \cmidrule(lr){8-9} \cmidrule(lr){10-10} \cmidrule(lr){11-12} \cmidrule(lr){13-14}
& MAE & MSE & MAE & MSE & MAE & MSE & MAE & MSE & Acc (\%) & MAE & MSE & MAE & MSE \\
\midrule
Autoformer    & 390.5 & 639.0 & 381.9 & 421.5 & 258.7 & 248.5 & 324.1 & 372.1 & 35.50 & 3.025 & 16.82 & 1751.6 & 6151.0 \\
DLinear       & 317.8 & 474.7 & 400.8 & 570.4 & 211.8 & 221.8 & 340.6 & 479.8 & 29.67 & 3.020 & 16.69 & 1632.4 & 5354.9 \\
TimesNet      & 376.7 & 628.7 & 435.1 & 556.4 & \second{210.7} & 200.4 & 336.6 & 446.2 & 33.42 & 3.016 & 16.78 & 1769.7 & 6000.1 \\
PatchTST      & 350.1 & 558.5 & 427.2 & 988.2 & 215.9 & 222.2 & 434.2 & 714.9 & 31.17 & 3.044 & 17.10 & 1713.8 & 5582.4 \\
Crossformer   & 300.9 & 386.2 & \second{318.1} & 369.0 & 211.2 & \second{188.0} & 323.2 & 400.9 & 31.33 & 2.948 & 15.96 & 1680.6 & 5496.1 \\
iTransformer  & 375.7 & 612.9 & 371.8 & 420.4 & 228.1 & 243.2 & 334.3 & 479.4 & 30.33 & 2.964 & 16.19 & 1652.7 & 5639.4 \\
TimeMixer     & 456.4 & 799.6 & 445.8 & 579.2 & 234.6 & 243.2 & 359.5 & 491.8 & 30.25 & 2.976 & 16.06 & 1669.2 & 5785.2 \\
\midrule
PromptCast    & 387.1 & 695.6 & 427.7 & 586.8 & 386.2 & 539.3 & 464.0 & 650.2 & 31.00 & 3.478 & 21.89 & 1931.0 & 7383.1 \\
One-Fits-All  & 314.5 & 415.5 & 565.5 & 2582.3 & 334.5 & 499.5 & 427.3 & 600.6 & 31.00 & 2.983 & 16.33 & \second{1612.1} & 5239.6 \\
Time-LLM      & 419.6 & 1171.0 & 446.3 & 1225.4 & 266.6 & 280.5 & 424.9 & 630.7 & 32.17 & 2.964 & 15.98 & 1631.3 & 5503.0 \\
CALF          & 294.1 & 420.4 & 361.1 & 425.6 & 233.4 & 213.8 & 334.7 & 442.7 & 32.08 & 2.989 & 16.33 & 1732.3 & 5806.6 \\
TimeCMA       & 324.9 & 483.6 & 393.0 & 747.7 & 237.6 & 213.0 & 348.7 & 418.3 & 32.50 & 2.953 & 15.85 & 1644.0 & 5423.1 \\
TimeCAP       & 302.5 & \second{376.2} & 323.6 & \second{343.0} & 251.5 & 230.4 & 354.9 & 460.0 & 30.25 & 3.055 & 17.05 & 1875.4 & 6381.0 \\
\midrule
TimesFM 3.0   & 428.7 & 760.6 & 450.6 & 642.6 & 300.0 & 334.7 & 417.3 & 590.6 & 34.67 & 3.105 & 17.76 & 1807.2 & 6386.3 \\
Chronos-2     & 407.2 & 727.9 & 433.0 & 579.1 & 273.2 & 298.0 & 396.1 & 548.3 & 28.92 & 3.237 & 19.64 & 1882.8 & 6925.6 \\
\midrule
\LLMev        & \second{287.9} & 377.5 & 372.8 & 466.8 & 234.5 & 211.1 & \second{296.6} & \second{299.9} & \second{36.92} & \second{2.590} & \second{12.69} & 1620.0 & \second{5189.7} \\
\midrule
\textbf{SEER} & \best{240.3} & \best{187.2} & \best{291.5} & \best{279.2} & \best{194.2} & \best{143.8} & \best{285.8} & \best{299.2} & \best{38.25} & \best{2.547} & \best{12.20} & \best{1473.3} & \best{4340.5} \\
\bottomrule
\end{tabular}%
}
\end{table*}

\begin{table*}[t]
\centering
\caption{\textbf{Pairwise LLM-as-a-judge (Grok-4.20-Reasoning)} evaluation on forecasting reasoning quality. Results demonstrate that \SEERfull{} significantly outperforms both the baseline (\LLMev{}, denoted as LLM-RAG) and the knowledge-ablated variant (\SEERnoK{}, denoted as SEER-NoK). Win / Tie / Loss denote the percentage (\%) of records where the first setting is better, tied, or loses. $\Delta$ represents the net win rate (Win $-$ Loss).}
\label{tab:judge-grok}
\small
\setlength{\tabcolsep}{4pt}
\renewcommand{\arraystretch}{1.08}
\resizebox{\textwidth}{!}{%
\begin{tabular}{@{}l cccc !{\vrule width 0.8pt} cccc !{\vrule width 0.8pt} cccc@{}}
\toprule
\textbf{Backbone} & \multicolumn{4}{c}{SEER-Full vs.\ LLM-RAG} & \multicolumn{4}{c}{SEER-Full vs.\ SEER-NoK} & \multicolumn{4}{c}{SEER-NoK vs.\ LLM-RAG} \\
\cmidrule(lr){2-5}\cmidrule(lr){6-9}\cmidrule(lr){10-13}
 & Win & Tie & Loss & $\Delta$ & Win & Tie & Loss & $\Delta$ & Win & Tie & Loss & $\Delta$ \\
\midrule
Claude Opus 4.8 & 71.8 & 20.3 & 7.9 & \cellcolor{green!20}\textbf{+64.0} & 67.6 & 26.4 & 6.1 & \cellcolor{green!20}\textbf{+61.5} & 32.2 & 44.1 & 23.6 & \cellcolor{green!20}\textbf{+8.6} \\
Claude Sonnet 4.6 & 54.8 & 30.4 & 14.8 & \cellcolor{green!20}\textbf{+40.0} & 44.4 & 37.3 & 18.3 & \cellcolor{green!20}\textbf{+26.2} & 40.5 & 41.1 & 18.4 & \cellcolor{green!20}\textbf{+22.0} \\
Gemini Flash 3.1 & 54.4 & 29.1 & 16.5 & \cellcolor{green!20}\textbf{+37.9} & 42.1 & 38.5 & 19.5 & \cellcolor{green!20}\textbf{+22.6} & 51.5 & 29.6 & 19.0 & \cellcolor{green!20}\textbf{+32.5} \\
Gemini Pro 3.1 & 41.3 & 30.5 & 28.3 & \cellcolor{green!20}\textbf{+13.0} & 41.0 & 32.7 & 26.2 & \cellcolor{green!20}\textbf{+14.8} & 33.3 & 33.0 & 33.8 & \cellcolor{red!15}\textbf{-0.5} \\
\midrule
\textbf{All backbones} & 55.6 & 27.6 & 16.9 & \cellcolor{green!20}\textbf{+38.7} & 48.8 & 33.7 & 17.5 & \cellcolor{green!20}\textbf{+31.3} & 38.7 & 36.6 & 24.7 & \cellcolor{green!20}\textbf{+14.1} \\
\bottomrule
\end{tabular}%
}
\end{table*}

\subsection{Ablation Study And Reasoning Evaluation}
\label{sec:exp-ablation}
To validate the effectiveness of the proposed self-evolving framework, \autoref{tab:abl-components} evaluates four LLM backbones across four forecasting configurations. These incremental settings systematically introduce initial events, memory-guided search and selection, and the causal knowledge of \SEERfull{}. The results yield three key takeaways. First, \SEERfull{} significantly outperforms basic event retrieval ``\LLMev{}'', yielding average improvements of +10.5\% for Claude 4.8 Opus and +11.4\% for Gemini 3.1 Flash. Second, SEER is model-agnostic; the complete architecture enhances predictive accuracy across all four evaluated backbones. Third, incorporating causal knowledge is critical for consistent gains. While memory-guided searching and filtering alone (\SEERnoK{}) yields improvements, integrating distilled causal knowledge systematically elevates forecasting accuracy. 

Additionally, we evaluate the reasoning quality using an LLM-as-a-judge approach adapted from \citet{tfrbench}, which conducts pairwise comparisons based on actual events alongside both models' textual reasoning and numerical predictions (withholding actual numerical outcomes to prevent outcome bias) across four dimensions: Domain Relevance, Event Relevance \& Plausibility, Logic-to-Number Consistency, and Analytical Depth (detailed in Appendix~\ref{app:llm-as-judeg-eval}). As evaluated by Grok-4.20-Reasoning (\autoref{tab:judge-grok}), \SEERfull{} demonstrates a clear advantage. Across the three pairwise comparisons, \SEERfull{} achieves the highest win rates, consistently outperforming both \SEERnoK{} and the \LLMev{} baseline.

\begin{table*}[t]
\centering
\caption{\textbf{Ablation study of SEER's components across multiple LLM backbones.} The best configuration per backbone is highlighted in \textbf{bold}. The final two columns, \textbf{vs.\ LLM} and \textbf{vs.\ $+$Events}, denote the average relative performance improvement across all seven tasks compared to the \LLMbare{} and \LLMev{} baselines. Results including 95\% confidence intervals are detailed in \autoref{tab:abl-components-ci} (Appendix~\ref{app:other-results}).}
\label{tab:abl-components}
\small
\setlength{\tabcolsep}{4pt}
\renewcommand{\arraystretch}{1.05}
\resizebox{\textwidth}{!}{%
\begin{tabular}{@{}llccccccccc@{}}
\toprule
\textbf{Backbone} & \textbf{Setting} & \textbf{Mem 6M} & \textbf{Mem 9M} & \textbf{SSD 6M} & \textbf{SSD 9M} & \textbf{Stock} & \textbf{Weather} & \textbf{Electricity} & \textbf{vs.\ LLM} & \textbf{vs.\ $+$Events} \\
 & & MAE & MAE & MAE & MAE & Acc.\ (\%) & MAE & MAE & $\Delta$ (\%) & $\Delta$ (\%) \\
\midrule
\multirow{4}{*}{\makecell[l]{Claude\\Opus 4.8}} & \LLMbare & 376.5 & 422.0 & 246.7 & 367.2 & 36.58 & 2.876 & 1642.2 & -- & -- \\
 & \LLMev & 287.9 & 372.8 & 234.5 & 296.6 & 36.92 & 2.590 & 1620.0 & +10.2 & -- \\
 & \SEERnoK & 303.7 & 370.1 & 218.6 & \textbf{276.6} & 37.00 & 2.589 & 1659.6 & +11.1 & +0.9 \\
 & \textbf{\SEERfull} & \textbf{240.3} & \textbf{291.5} & \textbf{194.2} & 285.8 & \textbf{38.25} & \textbf{2.547} & \textbf{1473.3} & \textbf{+19.5} & \textbf{+10.5} \\
\midrule
\multirow{4}{*}{\makecell[l]{Claude\\Sonnet 4.6}} & \LLMbare & 387.5 & 441.2 & 259.0 & 403.7 & 35.58 & 3.003 & 1734.3 & -- & -- \\
 & \LLMev & 296.0 & 401.3 & 250.8 & 323.8 & 37.25 & 2.651 & 1581.2 & +11.5 & -- \\
 & \SEERnoK & \textbf{286.9} & 390.4 & 275.2 & 324.4 & 37.58 & \textbf{2.632} & 1542.4 & +11.4 & -0.0 \\
 & \textbf{\SEERfull} & 302.6 & \textbf{351.3} & \textbf{207.4} & \textbf{304.8} & \textbf{38.00} & 2.645 & \textbf{1486.0} & \textbf{+17.1} & \textbf{+6.0} \\
\midrule
\multirow{4}{*}{\makecell[l]{Gemini\\Flash 3.1}} & \LLMbare & 409.0 & 439.9 & 269.9 & 394.5 & 34.00 & 2.808 & 1773.3 & -- & -- \\
 & \LLMev & 306.6 & 373.6 & 288.0 & 344.8 & 36.00 & 2.607 & 1818.1 & +8.1 & -- \\
 & \SEERnoK & \textbf{280.4} & 365.7 & 304.1 & 409.6 & 36.67 & 2.664 & \textbf{1751.9} & +6.6 & -1.5 \\
 & \textbf{\SEERfull} & 282.1 & \textbf{294.7} & \textbf{223.3} & \textbf{295.6} & \textbf{39.75} & \textbf{2.602} & 1753.7 & \textbf{+18.8} & \textbf{+11.4} \\
\midrule
\multirow{4}{*}{\makecell[l]{Gemini\\Pro 3.1}} & \LLMbare & 399.7 & 443.9 & 252.9 & 377.4 & 36.25 & 2.931 & 1611.0 & -- & -- \\
 & \LLMev & 313.2 & 382.6 & 230.1 & \textbf{295.1} & 37.08 & 2.555 & 1498.5 & +12.6 & -- \\
 & \SEERnoK & 310.2 & 380.9 & \textbf{204.0} & 301.5 & \textbf{37.42} & \textbf{2.544} & \textbf{1405.7} & +15.0 & +2.6 \\
 & \textbf{\SEERfull} & \textbf{275.8} & \textbf{283.6} & 229.9 & 303.7 & 36.83 & 2.593 & 1449.5 & \textbf{+17.0} & \textbf{+5.2} \\
\bottomrule
\end{tabular}%
}
\end{table*}

\subsection{Generalizability to Non-Time-Series Forecasting}
\label{sec:exp-non-tsf}

\begin{figure*}[t]
\centering
\includegraphics[width=\textwidth]{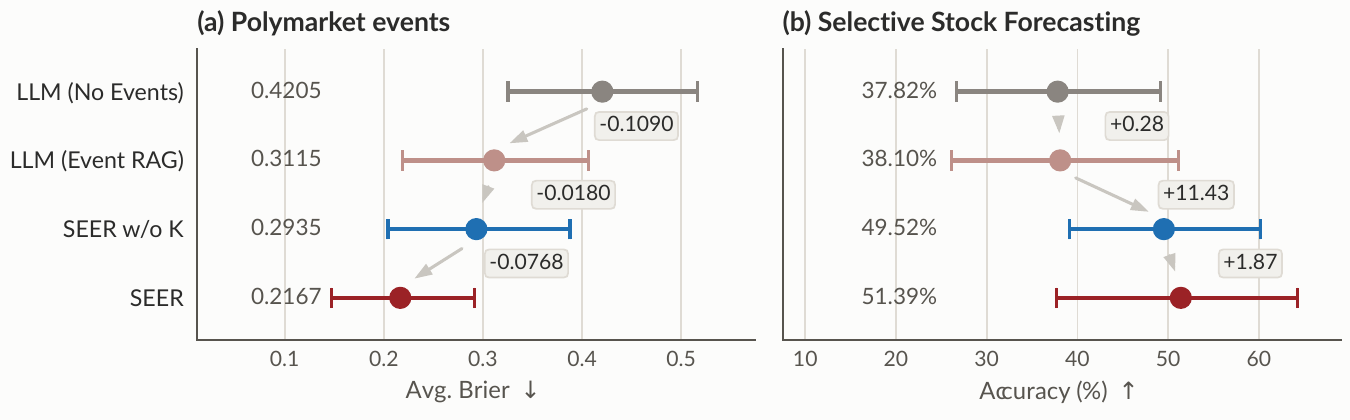}
\caption{\textbf{SEER transfers to non-time-series forecasting.} Performance comparison of different agent configurations on the Polymarket event forecsating task (evaluated by Brier Score) and the Selective Stock Forecasting task (evaluated by Accuracy). The progressive improvements highlight that SEER's evolving method, i.e., integrating both refined event retrieval and causal knowledge, consistently enhances predictive reasoning on event-driven tasks beyond standard numerical data.}
\label{fig:non-tsf-ladders}
\end{figure*}

To demonstrate the generality of our framework, we extend SEER beyond standard continuous time-series prediction to two broader domains: Polymarket event resolution and Selective Stock Forecasting. Although the forecasting targets of these tasks deviate from numerical trajectories, they share a underlying characteristic: their outcomes are driven by exogenous events and require complex causal reasoning. In the \textbf{Polymarket} task, the LLMs predict the evolving probabilities of discrete real-world outcomes (e.g., Fed interest rate decisions) prior to their final resolution. In the \textbf{Selective Stock Forecasting} task, the agent identifies and predicts the price direction for at most six equities from a daily pool of 50, which evaluates its ability to filter market noise. As shown in \autoref{fig:non-tsf-ladders}, SEER consistently outperforms baseline LLMs on both tasks, demonstrating that its evolving memory and knowledge generalize effectively to broader event-driven forecasting challenges.

\begin{figure*}[t]
\centering
\includegraphics[width=\textwidth]{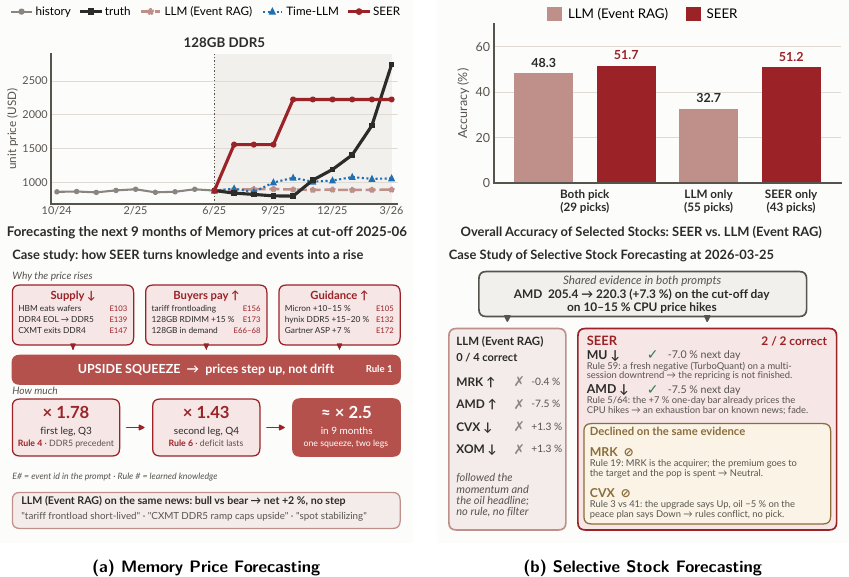}
\caption{\textbf{Interpretable forecasting of SEER.} (a) Memory price forecasting at the 2025-06-01 cut-off. In contrast to numerical models and the baseline \LLMev{} that miss the regime shift, SEER leverages distilled causal knowledge (e.g., UPSIDE SQUEEZE) to accurately predict the price surge. (b) Selective stock forecasting. (Top) For the 29 mutually selected stocks, \SEERfull{} and \LLMev{} achieve similar accuracies (51.7\% vs.\ 48.3\%). However, SEER selectively targets 43 different tickers that it evaluates as more predictable, on which it achieves a higher prediction accuracy than the baseline (51.2\% vs.\ 32.7\%), leading to an improved overall performance. (Bottom) An example illustrating how SEER leverages learned causal knowledge to filter out uncertain stocks. The details for the cases are provided in Figures~\ref{fig:128gb-ddr5-prompt} to \ref{fig:stockbatch-reasoning} in Appendix \ref{app:prompts-cases}.}
\label{fig:case-study}
\end{figure*}

\subsection{Explainable Forecasting via Causal Reasoning}
\label{sec:exp-analysis}

We qualitatively analyze SEER's underlying mechanisms to understand how its evolving environment drives performance improvements. \autoref{fig:case-study}(a) illustrates a Memory price forecasting scenario at the 2025-06 cut-off, a period marked by a sudden market regime shift. While numerical models and the baseline LLM (w/ Events) fail to anticipate this disruption, SEER successfully predicts the price surge by applying distilled causal knowledge. For instance, SEER utilizes the signal of ``HBM cannibalization'' (where expanding High-Bandwidth Memory production constrains standard Memory supply) alongside a ``step-wise multiplier decay'' to dynamically adjust the pricing magnitude. Beyond numerical value forecasting, \autoref{fig:case-study}(b) uncovers SEER's decision-making process in the Selective Stock Forecasting task. While SEER and the baseline LLM achieve comparable accuracy on their overlapping selections, SEER's overall performance gain stems from superior confidence calibration. Guided by distilled knowledge, SEER actively abstains from highly uncertain stocks, filtering out unpredictable market noise to maximize selection accuracy.

\section{Conclusion}
\label{sec:conclusion}
In this work, we introduce Self-Evolving Event Reasoning and Retrieval (SEER), a closed-loop framework designed to overcome inaccurate forecasting caused by insufficient input signals. By analyzing past correct and incorrect predictions, SEER optimizes two decoupled mechanisms: a reflective retrieval memory for supplementary event retrieval and noise filtering, and a persistent domain causal knowledge base to assist future forecasting. Extensive experiments demonstrate that SEER consistently outperforms state-of-the-art time series forecasting (TSF) methods. Furthermore, our results confirm that SEER generalizes effectively to highly volatile, event-driven tasks such as stock market and real-world event forecasting.



\bibliography{main}

\clearpage
\appendix

\section{Forecast Targets}
\label{app:targets}

This section lists the concrete entities behind each domain in Table~\ref{tab:data}.

\xhdr{Memory}
The paper targets are server memory modules in six capacities, 16, 32, 64, 96, 128 and 256\,GB, across two generations, DDR4 and DDR5. The number after the generation is the data rate in MT/s; the grades covered are DDR4-2400 and DDR4-3200, and DDR5-4800, 5600, 6400 and 7200. The public retail memory price are sourced from \cite{pangolyPriceTrends,pcpartpickerPriceTrends}, covering 8, 16, 32, 48, 64, 96, and 128\,GB kits in DDR4 and DDR5, at DDR4-3000, 3200, and 3600, and DDR5-4800, 5200, 5600, and 6000.

\xhdr{SSD}
Ten flash modules, from 32\,GB to 8\,TB: a 32\,GB MLC module, two 64\,GB TLC modules (one in the M.2~2280 form factor), 128\,GB and 256\,GB TLC modules, a 512\,GB module, and two 4\,TB and two 8\,TB modules whose NAND type is not specified in the part name. The data is from \cite{pangolyPriceTrends,pcpartpickerPriceTrends}.

\xhdr{Weather and Electricity}
Weather covers ten US cities: Boston, Chicago, Denver, Houston, Los Angeles, Miami, New York, San Francisco, Seattle and Washington, DC. Electricity covers three of them, Boston, New York and San Francisco, tracking the daily maximum and minimum load of the corresponding grid region.

\xhdr{Event Forecasting}
We adopt a subset of questions from ForecastBench~\cite{karger2024forecastbench}. Based on our chronological splitting strategy, the dataset comprises 100 training events and 46 testing events across three core domains: Tech, Economy, and Politics:
\begin{itemize}[leftmargin=*,itemsep=1pt,topsep=2pt]
    \item \textbf{Tech:} Predicting technological milestones and trends, such as the best LLM model and benchmark rankings.
    \item \textbf{Economy:} Forecasting macroeconomic indicators and financial events, such as Federal Reserve interest rate decisions.
    \item \textbf{Politics:} Anticipating global political shifts, such as international electoral outcomes and diplomatic developments.
\end{itemize}
Note that, we retrieve the specific questions, candidate answers, and final outcomes via PolyMarket API~\cite{polymarket2026api}.

\xhdr{Stocks}
The stock task covers 50 large US companies; By sector they are:
\begin{itemize}[leftmargin=*,itemsep=1pt,topsep=2pt]
  \item \emph{Technology and internet} (13): Meta (META), Apple (AAPL), Microsoft (MSFT), Alphabet (GOOGL), Amazon (AMZN), Tesla (TSLA), Netflix (NFLX), Disney (DIS), Adobe (ADBE), Salesforce (CRM), Oracle (ORCL), Uber (UBER), IBM (IBM).
  \item \emph{Semiconductors} (10): Nvidia (NVDA), AMD (AMD), Intel (INTC), TSMC (TSM), Broadcom (AVGO), Qualcomm (QCOM), Texas Instruments (TXN), Applied Materials (AMAT), Micron (MU), Lam Research (LRCX).
  \item \emph{Finance} (8): JPMorgan (JPM), Bank of America (BAC), Visa (V), Mastercard (MA), Berkshire Hathaway (BRK-B), Wells Fargo (WFC), Morgan Stanley (MS), Goldman Sachs (GS).
  \item \emph{Consumer} (7): Walmart (WMT), Costco (COST), Coca-Cola (KO), PepsiCo (PEP), Procter \& Gamble (PG), Home Depot (HD), McDonald's (MCD).
  \item \emph{Healthcare} (8): Johnson \& Johnson (JNJ), Pfizer (PFE), Eli Lilly (LLY), UnitedHealth (UNH), AbbVie (ABBV), Merck (MRK), Amgen (AMGN), Intuitive Surgical (ISRG).
  \item \emph{Industry and energy} (4): GE (GE), ExxonMobil (XOM), Chevron (CVX), Linde (LIN).
\end{itemize}
For each stock and cut-off the ground truth is one of three price change trend labels (Up, Neutral, Down), similar to~\citet{tan2026leaf}. The prompt of "Selective Stock Forecasting" can be found in \autoref{fig:stockbatch-prompt}.

\subsection{LLM-as-a-Judge Reasoning Evaluation}
\label{app:llm-as-judeg-eval}

To systematically evaluate the reasoning capabilities of LLMs in time-series forecasting, we employ \texttt{Grok-4.20-Reasoning} as an expert event-augmented forecasting judge. Different from standard accuracy metrics, this evaluation is formulated as a pairwise comparison between two candidate models, as detailed in \autoref{fig:judge-sxs-prompt}. To ensure the judge focuses on the \textit{quality, coherence, and plausibility} of the reasoning rather than being biased by the final outcome, actual ground-truth numerical values are withheld. For each evaluation instance, the judge is provided with: 1) the ground-truth events that occurred during the forecast horizon (specifically, those retrieved in the \LLMev{} setting), and 2) the reasoning text and numerical predictions generated by Model A and Model B.

The judge evaluates the models across four dimensions:

1) \textit{Domain Relevance}: Evaluates the appropriate use of domain-specific terminology (e.g., ``support levels'', ``seasonality'', ``macroeconomic headwinds'').

2) \textit{Event Relevance \& Plausibility}: Assesses whether the reasoning logically and causally links the provided ground-truth events to the predicted fluctuations, penalizing hallucinations or illogical connections.

3) \textit{Logic-to-Number Consistency}: Ensures alignment between the narrative plan and the model's \textit{own} numerical outputs (e.g., if the narrative predicts a ``sharp drop'', the predicted values must reflect that magnitude and direction).

4) \textit{Analytical Depth}: Measures whether the reasoning demonstrates an understanding of fundamental time-series dynamics (such as trend, volatility, and momentum) rather than providing superficial observations.

For each comparison, the judge outputs a JSON declaring the preferred model (Model A, Model B, or Tie) for each dimension and the overall preference, along with a textual justification citing specific criteria. The pairwise results in \autoref{tab:judge-grok} show that \SEERfull{} consistently outperforms both \SEERnoK{} and the static \LLMev{} baseline in generating logically coherent forecasting rationales.

\section{Prompts and Case Studies}
\label{app:prompts-cases}

\paragraph{Prompts.}
We provide all prompt templates utilized by SEER. Domain-specific variables are denoted as \{placeholders\}, with their instantiated values detailed in the respective figure captions.

\autoref{fig:seer-learn-mm} and~\ref{fig:seer-learn-know} illustrate the two-stage memory learning process. Part~A reflects on completed forecasts against realized outcomes to update the [SELECTION MEMORY] and [RETRIEVAL MEMORY]. Part~B, executed as an independent call using the same evidence, distills the [CAUSAL KNOWLEDGE]. Placeholders (e.g., \{name\}, \{reflect\_dynamics\}) ensure these templates adapt seamlessly across different domains.

\autoref{fig:seer-condense} outlines the three condensation prompts. These are triggered when a memory module exceeds 80\% of its character limit (12,000, 14,000, and 22,000 characters, respectively), rewriting the contents into a pruned and consolidated format.

\autoref{fig:seer-search-init} presents the initial event-search template. For the stock domain, this search executes daily, whereas a 7-day window is applied to other domains. This multi-round search process lists previously retrieved events (from round~2 onward) in the dashed block, retaining only those that pass LEAF's multi-agent fact-checking.

\autoref{fig:seer-search-missing} shows the expanded search template, which utilizes the learned retrieval memory to populate the \{missing\_info\} slot, specifically targeting blind spots identified during reflection.

Finally, \autoref{fig:seer-knowledge-ram}, \ref{fig:seer-knowledge-stockbatch}, and~\ref{fig:seer-knowledge-elec} present verbatim the causal knowledge acquired by SEER (Claude-Opus-4.8) by the end of training, precisely as inserted into the forecasting prompts. This includes 6 rules for memory pricing, 70 conditional IF-THEN rules for cross-stock batch forecasting (14 shown, maintaining original numbering), and 7 rules for city-level hourly electricity-load forecasting across three cities.

\paragraph{Case Studies.}
We further provide three complete execution transcripts to map the aforementioned templates to the qualitative case studies discussed in the main text.

\autoref{fig:64gb-ddr5-prompt} and~\ref{fig:64gb-ddr5-reasoning} detail the input and output for the 64GB DDR5 forecast (cut-off: 2025-05; \autoref{fig:seer_demo}(b)). The prompt integrates historical prices, retrieved events (filtered to those explicitly cited in the reasoning), and the six learned knowledge rules. In its response, SEER systematically applies the Rule-1 criteria to project an upward squeeze, establishing the finalized May price as the baseline. It then projects a $\times$1.780 multiplier at the next quarter boundary and a secondary $\times$1.43 multiplier at the subsequent one, ultimately outputting nine forecasted prices.

\autoref{fig:128gb-ddr5-prompt} and~\ref{fig:128gb-ddr5-reasoning} present a similar pipeline for the 128GB DDR5 forecast (cut-off: 2025-06; Figure~\ref{fig:case-study}(a)). Applying the same rule set, SEER leverages events [132], [156], [173], and [66--68] to construct a two-step forecast trajectory anchored to the finalized June price of 875.18.

\autoref{fig:stockbatch-prompt} and~\ref{fig:stockbatch-reasoning} illustrate the selective stock forecasting case (cut-off: 2026-03-25; \autoref{fig:case-study}(b)). The system prompt incorporates the learned causal rules, while the user prompt lists the 14-day historical closing prices and events for the 50-stock basket (displaying the data block for MU). In its response, SEER systematically rejects several candidates based on its rules: an acquirer following an M\&A-driven price spike (MRK, Rule~19), an oil major reacting to a single-headline reversal (CVX, Rules~3/41), and a parabolic single-session surge (AMD Up). SEER ultimately selects MU Down (Rule~59) and AMD Down (Rules~5/64). Both predictions successfully closed down in the subsequent session, whereas all four baseline predictions made on the same day were incorrect.

\section{Other Results}
\label{app:other-results}

\paragraph{Confidence Intervals.} 
To verify the statistical significance of our findings, we report 95\% confidence intervals (CIs) for both the main evaluations and ablation studies. To account for structural correlations, these CIs are computed via a cluster-bootstrap approach using 20,000 paired resamples, where all methods share the same draws. The resampling unit is defined by the forecast target: one memory product for Memory, one flash product for SSD,  a cut-off day for Stock, and a forecast window for Weather and Electricity (see Appendix \ref{app:targets}).

\autoref{tab:main-tsf-ci} and \autoref{tab:abl-components-ci} provide the CIs for the main forecasting results (\autoref{tab:main-tsf}) and component ablations (\autoref{tab:abl-components}), respectively. The intervals indicate a consistent performance advantage for SEER, with its estimated error ranges generally shifting lower than those of the baselines. Furthermore, the 95\% CIs for the average relative improvements (final two columns, \autoref{tab:abl-components-ci}) remain entirely positive across all evaluated LLM backbones, supporting the robustness of the dual-memory mechanisms.

\begin{table*}[t]
\centering
\renewcommand{\arraystretch}{2.1} 
\caption{95\% confidence intervals for every cell of Table~\ref{tab:main-tsf}. Each entry is the cluster-bootstrap interval (20{,}000 resamples; clusters are parts for Memory and SSD, cut-off days for Stock, forecast windows for Weather and Electricity) of the corresponding point estimate; all methods in a column are resampled with the same draws, so the intervals are paired. MSE for Memory, SSD and Electricity is scaled by $10^{-3}$. Bold and shading are carried over from Table~\ref{tab:main-tsf} (best and second-best point estimate).}
\label{tab:main-tsf-ci}
\small
\resizebox{\textwidth}{!}{%
\begin{tabular}{@{}l@{\hspace{4mm}} cc cc cc cc c cc cc @{}}
\toprule
\multirow{2}{*}{\textbf{Method}} 
& \multicolumn{2}{c}{\textbf{Memory 6M}} & \multicolumn{2}{c}{\textbf{Memory 9M}} 
& \multicolumn{2}{c}{\textbf{SSD 6M}} & \multicolumn{2}{c}{\textbf{SSD 9M}} 
& \textbf{Stock} 
& \multicolumn{2}{c}{\textbf{Weather}} & \multicolumn{2}{c}{\textbf{Electricity}} \\
\cmidrule(lr){2-3} \cmidrule(lr){4-5} \cmidrule(lr){6-7} \cmidrule(lr){8-9} \cmidrule(lr){10-10} \cmidrule(lr){11-12} \cmidrule(lr){13-14}
& MAE & MSE & MAE & MSE & MAE & MSE & MAE & MSE & Acc (\%) & MAE & MSE & MAE & MSE \\
\midrule
Autoformer    & [258.4, 548.5] & [244.3, 1279.3] & [258.3, 522.7] & [208.9, 675.9] & [89.9, 476.0] & [33.4, 579.8] & [124.8, 606.9] & [66.5, 936.9] & [31.67, 39.42] & [2.713, 3.357] & [13.47, 20.45] & [1392.1, 2150.3] & [3785.6, 8892.5] \\
DLinear       & [198.0, 467.0] & [158.3, 990.3] & [250.0, 575.5] & [245.4, 976.4] & [70.8, 395.1] & [29.4, 521.7] & [122.1, 658.4] & [72.9, 1209.7] & [25.92, 33.33] & [2.700, 3.352] & [13.38, 20.16] & [1325.1, 1978.5] & [3379.1, 7717.5] \\
TimesNet      & [245.6, 535.1] & [232.0, 1284.1] & [300.4, 587.4] & [270.6, 900.1] & \second{[72.2, 392.0]} & [26.9, 467.2] & [122.9, 649.6] & [67.7, 1128.0] & [30.17, 36.67] & [2.724, 3.324] & [13.56, 20.27] & [1490.4, 2077.2] & [4191.4, 8052.9] \\
PatchTST      & [223.6, 504.0] & [197.1, 1151.6] & [226.6, 735.0] & [181.0, 2636.1] & [70.7, 407.9] & [30.4, 517.6] & [155.3, 855.6] & [102.6, 1824.8] & [28.00, 34.33] & [2.732, 3.380] & [13.69, 20.78] & [1430.9, 2025.5] & [3843.4, 7631.5] \\
Crossformer   & [196.8, 434.9] & [143.2, 756.5] & \second{[206.4, 458.3]} & [146.9, 714.4] & [70.0, 399.1] & \second{[23.6, 440.3]} & [121.3, 611.8] & [70.7, 988.4] & [27.83, 34.83] & [2.646, 3.263] & [12.82, 19.29] & [1379.2, 2016.3] & [3584.9, 7771.8] \\
iTransformer  & [246.2, 530.1] & [225.7, 1256.3] & [249.6, 508.7] & [206.3, 677.8] & [75.1, 432.3] & [31.0, 576.3] & [121.9, 643.2] & [70.7, 1218.0] & [26.75, 34.25] & [2.663, 3.285] & [13.07, 19.53] & [1326.8, 2015.5] & [3477.8, 8205.2] \\
TimeMixer     & [308.4, 630.4] & [321.7, 1561.2] & [309.1, 600.8] & [282.5, 939.3] & [78.9, 439.0] & [32.7, 571.2] & [133.0, 688.6] & [78.3, 1235.8] & [27.25, 33.42] & [2.678, 3.285] & [13.03, 19.25] & [1328.1, 2038.5] & [3625.6, 8298.4] \\
\midrule
PromptCast    & [243.7, 559.7] & [248.5, 1417.8] & [275.8, 603.2] & [260.4, 992.2] & [126.4, 714.2] & [86.5, 1226.3] & [178.8, 843.7] & [134.1, 1531.5] & [28.42, 33.58] & [3.141, 3.816] & [17.99, 25.85] & [1573.3, 2339.6] & [4639.6, 10723.8] \\
One-Fits-All  & [204.9, 450.3] & [152.5, 849.8] & [234.9, 1138.8] & [188.5, 7696.6] & [106.1, 649.2] & [47.0, 1271.9] & [163.7, 792.3] & [121.8, 1413.8] & [27.25, 34.67] & [2.670, 3.312] & [13.14, 19.73] & \second{[1302.9, 1958.6]} & [3243.4, 7628.8] \\
Time-LLM      & [224.6, 701.4] & [220.5, 2985.6] & [225.1, 814.1] & [169.2, 3450.9] & [90.1, 502.1] & [33.2, 665.6] & [157.7, 827.6] & [96.6, 1613.3] & [27.67, 36.67] & [2.663, 3.278] & [12.87, 19.24] & [1303.7, 1998.9] & [3368.4, 8037.6] \\
CALF          & [187.1, 428.4] & [142.8, 888.7] & [239.6, 497.5] & [205.0, 692.7] & [80.4, 434.3] & [28.0, 498.5] & [125.8, 630.1] & [70.8, 1108.0] & [27.41, 36.92] & [2.687, 3.302] & [13.20, 19.68] & [1455.7, 2035.0] & [3980.8, 7921.9] \\
TimeCMA       & [207.4, 470.0] & [168.5, 1018.5] & [214.4, 667.9] & [153.2, 1955.7] & [82.4, 440.2] & [26.9, 498.6] & [132.1, 657.5] & [71.7, 1036.7] & [29.67, 35.42] & [2.667, 3.249] & [12.92, 18.94] & [1339.1, 1975.3] & [3528.8, 7555.0] \\
TimeCAP       & [199.9, 434.3] & \second{[145.1, 732.3]} & [211.7, 454.4] & \second{[159.6, 580.6]} & [86.1, 468.5] & [28.0, 540.4] & [133.7, 673.3] & [76.5, 1147.3] & [27.42, 33.17] & [2.734, 3.397] & [13.70, 20.65] & [1571.9, 2197.2] & [4483.1, 8480.7] \\
\midrule
TimesFM 3.0   & [278.1, 609.7] & [288.8, 1511.9] & [300.0, 626.4] & [295.9, 1078.6] & [100.1, 561.7] & [46.0, 778.7] & [152.4, 794.6] & [95.6, 1461.5] & [31.75, 37.58] & [2.766, 3.448] & [14.24, 21.42] & [1492.0, 2163.0] & [4192.5, 8976.7] \\
Chronos-2     & [264.5, 581.1] & [262.2, 1495.8] & [292.9, 594.3] & [272.8, 957.4] & [86.3, 521.9] & [38.3, 699.3] & [142.8, 752.7] & [89.1, 1357.6] & [26.17, 31.67] & [2.861, 3.624] & [15.63, 23.86] & [1559.7, 2241.6] & [4621.7, 9606.0] \\
\midrule
\LLMev        & \second{[187.3, 412.4]} & [132.0, 775.8] & [247.2, 517.5] & [215.9, 779.1] & [72.2, 456.2] & [22.6, 515.2] & \second{[114.3, 553.7]} & \second{[60.1, 741.0]} & \second{[34.08, 39.67]} & \second{[2.290, 2.889]} & \second{[9.90, 15.58]} & [1308.9, 1964.1] & \second{[3338.2, 7381.0]} \\
\midrule
\textbf{SEER} & \best{[158.3, 335.7]} & \best{[88.3, 311.2]} & \best{[191.8, 409.1]} & \best{[123.3, 480.2]} & \best{[69.1, 349.6]} & \best{[19.9, 330.0]} & \best{[111.5, 529.7]} & \best{[54.5, 741.6]} & \best{[34.00, 42.33]} & \best{[2.280, 2.810]} & \best{[9.74, 14.70]} & \best{[1217.1, 1768.8]} & \best{[2850.0, 6175.8]} \\
\bottomrule
\end{tabular}%
}\end{table*}

\begin{table*}[t]
\centering
\renewcommand{\arraystretch}{2}
\caption{\textbf{95\% confidence intervals for Table~\ref{tab:abl-components}.} Every cell shows the cluster-bootstrap interval (20{,}000 paired resamples) of the corresponding entry; the last two columns are the intervals of the average relative improvement over the seven tasks against \LLMbare{} and \LLMev{}, obtained by averaging the per-task paired bootstrap replicates. Bold follows the point estimates of Table~\ref{tab:abl-components}.}
\label{tab:abl-components-ci}
\small
\setlength{\tabcolsep}{4pt}
\resizebox{\textwidth}{!}{%
\begin{tabular}{@{}llccccccccc@{}}
\toprule
\textbf{Backbone} & \textbf{Setting} & \textbf{Mem 6M} & \textbf{Mem 9M} & \textbf{SSD 6M} & \textbf{SSD 9M} & \textbf{Stock} & \textbf{Weather} & \textbf{Electricity} & \textbf{vs.\ LLM} & \textbf{vs.\ $+$Events} \\
 & & MAE & MAE & MAE & MAE & Acc.\ (\%) & MAE & MAE & $\Delta$ (\%) & $\Delta$ (\%) \\
\midrule
\multirow{4}{*}{\makecell[l]{Claude\\Opus 4.8}} & \LLMbare & [240.2, 541.9] & [277.9, 589.9] & [81.1, 461.5] & [134.6, 693.7] & [33.08, 40.25] & [2.559, 3.212] & [1303.5, 2017.5] & -- & -- \\
 & \LLMev & [187.3, 412.4] & [247.2, 517.5] & [72.2, 456.2] & [114.3, 553.7] & [34.08, 39.67] & [2.290, 2.889] & [1308.9, 1964.1] & [+6.2, +14.8] & -- \\
 & \SEERnoK & [191.8, 444.9] & [245.6, 513.0] & [72.6, 411.9] & \textbf{[102.7, 531.5]} & [33.33, 40.75] & [2.315, 2.864] & [1371.8, 1976.6] & [+6.9, +15.9] & [-1.6, +3.7] \\
 & \textbf{\SEERfull} & \textbf{[158.3, 335.7]} & \textbf{[191.8, 409.1]} & \textbf{[69.1, 349.6]} & [111.5, 529.7] & \textbf{[34.00, 42.33]} & \textbf{[2.280, 2.810]} & \textbf{[1217.1, 1768.8]} & \textbf{[+14.0, +26.1]} & \textbf{[+6.4, +15.9]} \\
\midrule
\multirow{4}{*}{\makecell[l]{Claude\\Sonnet 4.6}} & \LLMbare & [254.6, 554.1] & [292.3, 615.4] & [86.5, 488.3] & [152.5, 757.7] & [32.58, 38.50] & [2.677, 3.367] & [1392.7, 2122.0] & -- & -- \\
 & \LLMev & [196.4, 427.5] & [266.2, 560.0] & [87.5, 458.1] & [128.2, 599.4] & [34.00, 40.83] & [2.393, 2.911] & [1283.2, 1912.7] & [+7.6, +16.1] & -- \\
 & \SEERnoK & \textbf{[191.2, 409.9]} & [261.2, 542.1] & [89.4, 533.2] & [121.6, 607.0] & [34.42, 40.67] & \textbf{[2.370, 2.890]} & [1243.2, 1871.7] & [+7.5, +15.6] & [-3.9, +3.0] \\
 & \textbf{\SEERfull} & [199.8, 430.0] & \textbf{[231.0, 493.1]} & \textbf{[73.0, 384.9]} & \textbf{[119.6, 572.4]} & \textbf{[34.17, 41.75]} & [2.380, 2.914] & \textbf{[1219.1, 1775.0]} & \textbf{[+12.6, +22.5]} & \textbf{[+2.8, +9.4]} \\
\midrule
\multirow{4}{*}{\makecell[l]{Gemini\\Flash 3.1}} & \LLMbare & [267.3, 583.2] & [292.9, 609.3] & [92.2, 515.7] & [145.9, 751.0] & [30.00, 38.17] & [2.521, 3.122] & [1430.5, 2169.4] & -- & -- \\
 & \LLMev & [195.3, 447.2] & [242.6, 521.9] & [112.5, 497.1] & [133.4, 628.4] & [32.58, 39.42] & [2.336, 2.882] & [1484.3, 2199.3] & [+4.3, +12.8] & -- \\
 & \SEERnoK & \textbf{[182.9, 404.6]} & [232.1, 518.4] & [115.3, 529.3] & [144.9, 727.0] & [33.42, 39.92] & [2.376, 2.955] & \textbf{[1419.0, 2132.1]} & [+0.2, +13.4] & [-6.8, +3.2] \\
 & \textbf{\SEERfull} & [192.6, 396.6] & \textbf{[204.4, 396.8]} & \textbf{[84.0, 409.1]} & \textbf{[110.3, 573.9]} & \textbf{[36.92, 42.67]} & \textbf{[2.329, 2.874]} & [1457.6, 2085.9] & \textbf{[+13.7, +24.8]} & \textbf{[+7.3, +15.4]} \\
\midrule
\multirow{4}{*}{\makecell[l]{Gemini\\Pro 3.1}} & \LLMbare & [257.1, 573.0] & [294.6, 617.0] & [86.6, 477.0] & [141.9, 716.6] & [33.00, 39.50] & [2.608, 3.285] & [1305.0, 1968.1] & -- & -- \\
 & \LLMev & [201.0, 450.7] & [254.8, 532.4] & [80.7, 424.0] & \textbf{[112.8, 560.5]} & [34.33, 39.83] & [2.285, 2.822] & [1250.7, 1795.9] & [+8.5, +17.6] & -- \\
 & \SEERnoK & [202.8, 441.0] & [252.6, 526.9] & \textbf{[75.5, 371.2]} & [117.7, 562.3] & \textbf{[33.83, 40.92]} & \textbf{[2.295, 2.787]} & \textbf{[1178.5, 1681.4]} & [+10.1, +21.1] & [-1.0, +6.3] \\
 & \textbf{\SEERfull} & \textbf{[182.2, 386.3]} & \textbf{[189.4, 395.4]} & [77.1, 437.2] & [113.9, 581.4] & [34.17, 39.83] & [2.356, 2.819] & [1193.5, 1754.0] & \textbf{[+12.6, +22.1]} & \textbf{[+1.2, +9.2]} \\
\bottomrule
\end{tabular}%
}\end{table*}

\clearpage

\lstset{basicstyle=\ttfamily, columns=flexible, breaklines=true, mathescape=true}

\tcbset{
  aibox/.style={
    width=402.pt,
    top=10pt,
    colback=white,
    colframe=black,
    colbacktitle=black,
    enhanced,
    center,
    attach boxed title to top left={yshift=-0.1in,xshift=0.15in},
    boxed title style={boxrule=0pt,colframe=white,},
  }
}
\newtcolorbox{AIbox}[2][]{aibox,title=#2,#1}

\begingroup
\tcbset{breakable,enhanced jigsaw}
\tiny
\setlength{\parindent}{0pt}
\begin{AIbox}{{System Prompt For Memory Price Forecasting}}
**Task:** \\
You will be provided with a RAM product's recent monthly **unit prices** and a list of news events (or N/A) affecting that part. Your task is to forecast the **exact monthly unit prices** (in USD) for the next 9 consecutive months following the cut-off month. \\
 \\
**Notes:** \\
* Prices are observed monthly; each value represents the unit price for that month. \\
* The prediction targets are the next 9 specific months listed below; you must emit one price per target month, in chronological order. \\
* Prediction Knowledge is the historical forecasting experience; please refer to it when making predictions. \\
\end{AIbox}
\begin{AIbox}{{User Prompt For 64GB DDR5 at Cut-off 2025-05 (19 of 68 events shown, 6 knowledge rules)}}
**Target Part:** 64GB-DDR5 \\
**Cut-off Month:** 2025-05 \\
**Prediction Horizon:** Next 9 months \\
**Target Months (in order):** 2025-06, 2025-07, 2025-08, 2025-09, 2025-10, 2025-11, 2025-12, 2026-01, 2026-02 \\
 \\
**A. Historical Monthly Prices (2024-09 to 2025-05)** \\
227.26, 228.41, 224.48, 220.41, 224.68, 217.35, 217.03, 211.12, 229.86 \\
 \\
**B. Event Intelligence** \\
{}[3] 2024-10-29: Late-October market dynamics reports highlighted that major memory suppliers (Samsung, SK Hynix, Micron) are maintaining a firm stance on DDR5 price hikes specifically because High Bandwidth Memory (HBM) production is displacing conventional DRAM capacity. \\
... \\
{}[41] 2024-12-16: SK Hynix Chairman Chey Tae-won confirmed the company's 2025 HBM capacity is fully booked and announced the acceleration of the M15X fab staffing to advance HBM4 supply by six months, driven by an explicit request from Nvidia CEO Jensen Huang. \\
{}[42] 2024-12-18: Micron Technology reported Q1 FY2025 financial results, announcing that data center revenue surpassed 50\% of total revenue for the first time, growing over 400\% year-over-year, and that sequential DRAM prices increased in the high-single-digit percentage range. \\
{}[43] 2024-12-18: Micron Technology confirmed that its High Bandwidth Memory (HBM) production is completely sold out for calendar year 2025 with locked-in pricing, and raised its 2025 HBM Total Addressable Market (TAM) estimate to over \$30 billion. \\
... \\
{}[74] 2025-01-22: TrendForce announced that major memory makers, including Micron, Samsung, and SK Hynix, are systematically implementing production cuts by lowering fab utilization rates and delaying manufacturing process upgrades. \\
... \\
{}[92] 2025-02-17: SK hynix accelerated the completion of its M15X fab in Cheongju to November 2025, announcing plans to deploy engineers in March to kick-start operations that will increase the company's HBM capacity by 20\% to 30\% from current levels. \\
{}[93] 2025-02-18: Micron forecast that overall DRAM prices will rise starting in Q2 2025, as manufacturers increasingly prioritize High-Bandwidth Memory (HBM) production for AI servers at the expense of standard DDR5 DRAM. \\
{}[94] 2025-02-18: According to DIGITIMES, the three major DRAM manufacturers---Samsung Electronics, SK Hynix, and Micron---announced plans to discontinue DDR3 and DDR4 production within 2025 to concentrate all production resources on DDR5 and High Bandwidth Memory (HBM). \\
... \\
{}[114] 2025-03-25: TrendForce issued Q2 2025 guidance projecting that average DRAM pricing including HBM will rise 3-8\% QoQ driven by increasing HBM3e 12hi shipments, while explicitly noting that SK hynix is prioritizing its wafer capacity for server and mobile DRAM, thereby limiting conventional DDR5 supply. \\
{}[115] 2025-03-25: DRAMeXchange (TrendForce) reported that Server DRAM DDR5 prices officially bottomed out for Q2 2025, keeping overall pricing flat QoQ despite narrowed DDR4 declines, driven by HBM demand pressure and production shift bottlenecks. \\
{}[116] 2025-03-31: Micron Technology officially confirmed plans to raise DRAM and NAND flash prices through 2025 and 2026, citing severe supply constraints and surging demand from AI data centers. \\
... \\
{}[124] 2025-04-17: TrendForce reported that the implementation of U.S. reciprocal tariffs prompted memory buyers to adopt a defensive stance, actively raising DRAM and NAND Flash inventory levels during a 90-day grace period and expanding anticipated Q2 contract price increases. \\
{}[125] 2025-04-17: TrendForce Senior Research Vice President Avril Wu reported that the U.S. 'reciprocal tariffs' implemented on April 9th caused memory buyers to actively stockpile DRAM inventory during the 90-day grace period. \\
... \\
{}[134] 2025-04-22: Samsung scheduled the cessation of its DDR4 memory production for early June 2025, strategically reallocating its semiconductor manufacturing capacity toward higher-margin DDR5 and High Bandwidth Memory (HBM) products. \\
... \\
{}[136] 2025-04-23: TrendForce reported that DRAM suppliers raised spot prices by 8-10\% as OEMs aggressively increased inventory levels to build defensive buffers against supply risks ahead of U.S. reciprocal tariff implementations. \\
... \\
{}[141] 2025-05-05: TrendForce reported that SK hynix increased its consumer DRAM chip and die prices by 12\%, joining earlier hikes by Micron and Samsung. \\
... \\
{}[144] 2025-05-12: Samsung finalized a new DRAM supply contract with major customers, actively raising contract prices for DDR4 by 20\% and DDR5 by 5\% in response to aggressive pre-tariff stockpiling. \\
{}[150] 2025-05-21: TrendForce reported that major memory suppliers, including Samsung, SK hynix, and Micron, have announced End-of-Life (EOL) plans for DDR4 products with final shipments scheduled for early 2026, driving DDR4 contract prices up by over 20\% in Q2. \\
{}[153] 2025-05-27: Under a Chinese government directive, ChangXin Memory Technologies (CXMT) executed a dramatic production shift away from legacy DDR4 to manufacturing DDR5 memory. \\
... \\
 \\
**C. Prediction Knowledge** \\
1. **Regime Classification \& Foundational Drivers** \\
Before forecasting any SKU, explicitly classify the regime (Squeeze vs. Glut) using a 4-point checklist to avoid catastrophic sign errors: \\
1. Is supply expanding (rising utilization/wafer output) or contracting? \\
2. Are downstream buyers accepting quoted price hikes or resisting/threatening non-purchase? \\
3. Is supplier guidance narrowing (loosening) or widening (tightening)? \\
4. Is the specific SKU trailing-edge (glut-prone) or leading-edge? \\
*Verdict:* If supply is contracting and buyers are accepting hikes (or ``sold out'' announcements occur), it is an UPSIDE squeeze. If supply is expanding and buyers resist, it is a DOWNSIDE cycle top. \\
*Corollaries:* \\
- CEO/manufacturer forward pricing language (``prices rising throughout the year'') strongly overrides single-quarter analyst guidance. \\
- In a structural supply-deficit, price is set at the margin by supply scarcity; weak conventional demand does NOT cap prices. \\
- Spot market weakness (e.g., spot DDR4 drifting down) is transient retail noise; ignore it completely when forecasting CONTRACT SKUs. \\
\S \\
2. **Execution Hierarchy \& Math Discipline** \\
Forecast error is dominated by a strict precision hierarchy: (1) Correct regime sign, (2) Correct base selection, (3) Correct step timing (Q-boundary placement), and (4) Multiplier magnitude. \\
- *Target Math:* Compute the target strictly as: \texttt{Target = Correct\_Base $\times$ Precedent\_Multiplier}. \\
- *Rounding Discipline:* Carry one extra significant figure through the multiplication (e.g., Base $\times$ 1.780) before rounding the final target to two decimals. This shaves residual basis points and ensures cent-exact accuracy. \\
- *Error Propagation:* Base errors propagate multiplicatively. Base-inflation compounds magnitude misses: overshooting the pre-boundary base while simultaneously under-applying the boundary multiplier produces a catastrophic ``double miss.'' Always anchor the structural leg to the correct SETTLED base. \\
- *Commitment:* Once the base, timing, and sign are locked, apply the SKU-specific realized multiplier with complete commitment. Do not artificially hedge, second-guess, or fine-tune to chase sub-0.3\% precision. \\
\S \\
3. **Base Calibration \& Handling Guidance** \\
- *Settled Base:* If historical prints show the current quarter settled, carry that exact value forward flat to two decimals. Do NOT inflate it mid-quarter. \\
- *Synthesized Base from Guidance:* \\
  - *Modest Guidance (e.g., 3-8\%):* Settlements tend to clear near the arithmetic MIDPOINT. \\
  - *Demand Caveat:* If suppliers/analysts explicitly note demand ``did not materialize as anticipated'' or guidance was ``narrowed,'' the settlement clears in the LOWER HALF of the range (e.g., \textasciitilde{}4\%), NOT the absolute minimum (\textasciitilde{}3\%). Only push to the true floor when demand language is unambiguously weak AND supply is loosening. \\
- *Aggregate vs. SKU:* Do NOT convert an industry-aggregate contract-price \% (e.g., TrendForce ``13-18\% Q2 DRAM increase'') into a per-SKU price step. Aggregate figures blend many parts; the SKU's OWN historical prints already encode its settled move. Fabricating an intra-quarter lift to reconcile a SKU print with an aggregate headline is a critical failure mode. \\
\S \\
4. **Structural Multipliers \& Wafer Cannibalization** \\
Wafer reallocation to premium products creates violent, discontinuous price spikes. HBM consumes leading-edge wafers at a 3:1 bit ratio, starving both leading-edge and trailing-edge SKUs. Trailing-edge starvation often produces equal or greater price explosions due to severe capacity cuts and inelastic demand. \\
*First Full Structural Leg Multipliers:* Match the target SKU on node, capacity, and form-factor. \\
- Leading-edge DDR5 Server RDIMM: \textasciitilde{}1.78x (Realized exactly 1.780x. Do not use the older \textasciitilde{}1.807x ladder value for this SKU). \\
- Trailing-edge DDR4 32GB Module: \textasciitilde{}1.845x \\
- Scarcest Trailing-edge DDR4 64GB RDIMM: \textasciitilde{}1.880x \\
*Shock Amplification:* When a structural step is driven by a one-time physical shock (earthquake, fab outage) combined with a \textgreater{}20-25\% manufacturer hike notice, the realized settlement (e.g., \textgreater{}78\% jump) far EXCEEDS the nominal hike-notice \%. The hike notice acts as a floor, and the physical shock amplifies it. \\
\S \\
5. **Shape, Timing, and Plateau Discipline (Step-Then-Plateau)** \\
Contract-priced RAM SKUs move in discontinuous steps. The default shape is a flat base, ONE large discrete step, and then perfectly FLAT again. \\
- *Current Quarter Hold:* Hold the CURRENT quarter perfectly flat at the settled print. Even if your reasoning trace explicitly identifies an accelerating deficit ramping into the next quarter, acceleration expresses itself at the NEXT Q-boundary, not by retroactively inflating the already-settled current quarter. \\
- *Timing the Step:* Place the full structural step exactly at the FIRST month of the NEXT quarter (Q-boundary). The ABSENCE of published next-quarter analyst guidance is NEVER a valid reason to hold flat into an identified accelerating deficit. Structural squeezes reprice on physical scarcity, not on whether analysts have quantified a number. \\
- *Plateau Discipline:* Once a contract reprices, it holds perfectly flat. Never sprinkle small monthly gains (``marginal firmness'') onto a plateau, never apply a reflexive decay tick, and ignore sub-0.2\% micro-wobble within a settled quarter. One-time discrete physical shocks (earthquakes) ALWAYS produce a single step followed immediately by a flat plateau. \\
\S \\
6. **Multi-Leg Squeezes \& Leg-Counting Discipline** \\
Structural squeezes can arrive as multiple sequential legs, but you must strictly categorize prior steps to assign the correct multiplier. \\
- *Leg-Counting:* Modest single-digit intra-quarter steps (+3-8\%) do NOT count as ``structural legs.'' If the prior quarter merely settled at a mild step, the next Q-boundary jump is a FIRST full structural leg and clears at the \textasciitilde{}1.78x - 1.88x floor (See Knowledge 4). \\
- *Second Leg Multiplier:* If the base already contains a violent structural spike (\textasciitilde{}1.8x), the NEXT structural leg at a later Q-boundary clears at the upper edge of the secondary band: \textasciitilde{}1.43x. \\
- *Within-Quarter Staircase (Rare Exception):* Default shape is always Step-Then-Plateau. A high accelerant count of long-horizon narrative items (new fab CapEx, ``sold out through 2025,'' GPU roadmaps) supports the FIRST leg and the plateau level---they do NOT generate fresh monthly steps. ONLY real-time worsening of a deficit within the quarter (rising HBM capacity share prints, sequential real-time hike notices) justifies a second within-quarter leg (\textasciitilde{}1.18x month-over-month). When uncertain, strongly prefer the plateau; over-forecasting multiple legs based on narrative backdrop is a primary failure mode. \\
 \\
**TASK:** \\
Predict the **exact monthly unit prices** (USD, floats) for these target months IN ORDER: 2025-06, 2025-07, 2025-08, 2025-09, 2025-10, 2025-11, 2025-12, 2026-01, 2026-02. \\
 \\
**STRICT CONSTRAINTS:** \\
1. **COMPREHENSIVE ANALYSIS:** Provide an in-depth Analysis of event-driven price impacts before emitting numbers. \\
2. **FORMAT:** Inside \texttt{\texttt{<prediction>}} tags, output EXACTLY 9 numeric values separated by commas, in chronological order matching the target months. Just the comma-separated floats. \\
3. **WRAPPER:** You MUST wrap the comma-separated prices inside \texttt{\texttt{<prediction>}} tags. \\
 \\
**Output:** \\
{}[COMPREHENSIVE Analysis] \\
\texttt{<prediction>}price\_1, price\_2, price\_3, price\_4\texttt{</prediction>}
\end{AIbox}
\normalsize
\captionof{figure}{The complete SEER input for 64GB DDR5 at the 2025-05 cut-off (\autoref{fig:seer_demo}(b)). The system and user prompts utilize the memory price forecasting templates, which consist of the task description and historical prices (Section A), the retrieved events (Section B), and the causal forecasting knowledge distilled from past predictions (Section C). Due to space limitations, certain events are omitted; only those explicitly referenced during the reasoning process (\autoref{fig:64gb-ddr5-reasoning}) are retained. In the ablation studies, when events or knowledge are excluded, their respective sections are marked as ``N/A''.}
\label{fig:64gb-ddr5-prompt}
\endgroup

\ifcsname AIbox\endcsname\else
  \newtcolorbox{AIbox}[2][]{colback=gray!5!white,colframe=black!60!white,fonttitle=\bfseries,
    title=#2,boxrule=0.5pt,arc=1.5pt,left=3pt,right=3pt,top=2pt,bottom=2pt,#1}
\fi
\begin{figure*}[t]
\centering
\tiny
\begin{AIbox}{{SEER's Reasoning and Forecast}}
\# COMPREHENSIVE ANALYSIS \\
 \\
\#\# Regime Classification (4-Point Checklist) \\
 \\
1. **Supply expanding or contracting?** CONTRACTING for the effective DDR5 supply. HBM is cannibalizing leading-edge wafers at a 3:1 bit ratio. Samsung, SK Hynix, and Micron are all phasing out DDR4 and prioritizing HBM/premium DDR5. SK Hynix explicitly limiting conventional DDR5 supply (Event 114). \\
2. **Buyers accepting hikes?** YES. Events 124/125/136/144 confirm aggressive pre-tariff stockpiling; buyers frontloaded shipments and accepted 5-20\% hikes. Samsung raised DDR5 contract prices 5\% (Event 144); SK Hynix raised consumer DRAM 12\% (Event 141). \\
3. **Guidance narrowing (tightening)?** TIGHTENING. Micron confirmed DRAM price hikes through 2025 AND 2026 (Event 116). TrendForce Q2 guidance +3-8\% QoQ with server DRAM bottomed out (Event 114/115). \\
4. **Trailing- or leading-edge SKU?** 64GB DDR5 is a leading-edge server/high-density module. \\
 \\
**VERDICT: UPSIDE SQUEEZE.** Supply contracting + buyers accepting hikes + CEO forward-pricing language (``prices rising throughout 2025-2026''). This is a structural supply-deficit regime. \\
 \\
\#\# Base Calibration \\
 \\
The historical series shows the SKU drifting down through late 2024 (glut tail) then reversing sharply: 211.12 (Apr) $\rightarrow$ 229.86 (May). The May print of **229.86** reflects the Q2 repricing beginning as pre-tariff stockpiling and contract hikes hit. \\
 \\
Q2 2025 (Apr-Jun) is the current quarter. Per Knowledge 5 (Current Quarter Hold), once the quarter has settled I hold it flat. May at 229.86 represents the Q2 settlement level for this SKU. June should hold flat to the settled Q2 print. \\
 \\
\#\# Structural Step Timing \\
 \\
The next Q-boundary is **Q3 2025 = July 2025**. Per Knowledge 5, the full structural step is placed at the first month of the next quarter (July). \\
 \\
For a leading-edge DDR5 Server RDIMM-class SKU, Knowledge 4 specifies the First Full Structural Leg Multiplier = **1.78x**. \\
 \\
**Leg-counting check (Knowledge 6):** The move from 211.12 $\rightarrow$ 229.86 (\textasciitilde{}8.9\%) is a modest intra-quarter step, NOT a structural leg. Therefore the July Q-boundary jump qualifies as the FIRST full structural leg $\rightarrow$ apply \textasciitilde{}1.78x. \\
 \\
\#\# Target Math \\
 \\
- **Base = 229.86** (settled Q2 print) \\
- **July (Q3 first month):** 229.86 $\times$ 1.780 = 409.15 \\
 \\
Per Knowledge 5 (Plateau Discipline), after the structural step, hold perfectly flat through Q3 (Jul, Aug, Sep). \\
 \\
\#\# Q4 2025 Boundary (October) \\
 \\
Next Q-boundary = October. The base now contains a violent structural spike (\textasciitilde{}1.78x), so per Knowledge 6, the NEXT structural leg clears at the secondary band \textasciitilde{}**1.43x**. \\
 \\
Structural drivers continuing into Q4: CXMT government-directed shift entirely to DDR5 (removes cheap alternative supply), DDR4 EOL final shipments early 2026 (Event 150), Micron pricing through 2026, HBM capacity ramps (SK Hynix M15X Nov 2025). Deficit persists and intensifies. \\
 \\
- **October:** 409.15 $\times$ 1.430 = 585.08 \\
 \\
Hold flat through Q4 (Oct, Nov, Dec) per plateau discipline. \\
 \\
\#\# Q1 2026 Boundary (January) \\
 \\
Next Q-boundary = January 2026. Deficit continues but a third sequential \textasciitilde{}1.43x leg would be over-forecasting. Structural narrative supports continued firmness but the acute spike drivers (pre-tariff stockpiling) are exhausting. Per Knowledge 6, avoid over-forecasting multiple legs. However, DDR4 final EOL shipments hit early 2026 (Event 150), removing the last legacy substitute --- this justifies a moderate continued structural leg but well below the secondary band. I apply a tempered step (\textasciitilde{}1.10x) reflecting continued but decelerating deficit repricing. \\
 \\
- **January 2026:** 585.08 $\times$ 1.10 = 643.59 \\
 \\
Hold flat through Q1 (Jan, Feb). \\
 \\
\#\# Final Forecast \\
 \\
| Month | Price | \\
|-------|-------| \\
| 2025-06 | 229.86 (Q2 hold) | \\
| 2025-07 | 409.15 (structural leg 1.78x) | \\
| 2025-08 | 409.15 (plateau) | \\
| 2025-09 | 409.15 (plateau) | \\
| 2025-10 | 585.08 (structural leg 1.43x) | \\
| 2025-11 | 585.08 (plateau) | \\
| 2025-12 | 585.08 (plateau) | \\
| 2026-01 | 643.59 (moderate leg 1.10x) | \\
| 2026-02 | 643.59 (plateau) | \\
 \\
\texttt{<prediction>}229.86, 409.15, 409.15, 409.15, 585.08, 585.08, 585.08, 643.59, 643.59\texttt{</prediction>}
\end{AIbox}
\caption{SEER's reasoning and forecast for the prompt in Figure~\ref{fig:64gb-ddr5-prompt}.  It walks the Rule-1 regime checklist to an upside squeeze, citing the events by number,
freezes the settled May print as the base (Rule~3, 5), places one $\times$1.780 step at the next quarter
boundary (Rule~4, 5) and a $\times$1.43 second leg at the following one (Rule~6), then emits nine prices
inside \texttt{<prediction>} tags.}
\label{fig:64gb-ddr5-reasoning}
\end{figure*}

\clearpage

\begingroup
\tcbset{breakable,enhanced jigsaw}
\fontsize{7}{7.2}\selectfont 
\setlength{\parindent}{0pt}

\begin{AIbox}{{SEER Memory Learning (Part A) --- System Prompt: [SELECTION MEMORY] + [RETRIEVAL MEMORY]}}
**TASK:** Optimize the [SELECTION MEMORY] and [RETRIEVAL MEMORY] modules by comparing the LLM's monthly **\{name\} unit-price** forecast against the realized ground-truth prices. \\
([SELECTION MEMORY] = event-SELECTION patterns for picking high-impact, unpriced supply/demand and structural catalysts from the event pool. [RETRIEVAL MEMORY] = event-SUPPLEMENTATION reminders --- predictive signal types / data sources absent from the pool.) \\
* **Reflect:** Compare the predicted monthly unit prices and the LLM's reasoning trace point-by-point against the ground-truth price trajectory. Identify which events / mechanisms drove correct vs. incorrect movements, and what magnitude or direction errors say about overlooked structural dynamics (e.g., \{reflect\_dynamics\}). \\
* **Context:** Everything provided below for this reflection --- the entire Local Event Pool, the forecasting conditions, the prediction results (forecasting process + ground-truth prices), and the current states of [SELECTION MEMORY] and [RETRIEVAL MEMORY]. \\
* **Update [SELECTION MEMORY] (event SELECTION):** Helps pick high-impact events and filter noise from the local event pool. Record general criteria for *which types* of events move \{name\} prices most --- transferable selection patterns, not this SKU's or month's specifics. \\
* **Update [RETRIEVAL MEMORY] (event SUPPLEMENTATION):** Reminds the system to seek absent, forecast-moving events for the local event pool. Record broad categories, recurring blind spots, or patterns of information to look for next time (e.g., \{missing\_signals\}) --- not a single missing fact.
\end{AIbox}
\begin{AIbox}{{SEER Memory Learning (Part A) --- User Prompt}}
(1) Current Event Pool: \\
\{event\_list\_pool\} \\
 \\
(2) Prediction Conditions: \\
**Target Part:** \{company\_name\} \\
**Cut-off Month:** \{fore\_cut\_off\_date\} \\
**Prediction Horizon:** Next \{prediction\_horizon\} months unit price \\
 \\
**A. Historical Monthly Unit Prices (From \{start\_trading\_date\} to \{fore\_cut\_off\_date\})** \\
\{his\_prices\} \\
 \\
**B. Events Selected** \\
\{event\_list\_selected\} \\
 \\
(3) Prediction Process \& Label: \\
\{llm\_forecasting\_resp\} \\
 \\
**Ground-Truth Monthly Unit Prices:** \{label\} \\
 \\
(4) [SECTOR] Information: \\
\{sector\} \\
 \\
(5) Current [SELECTION MEMORY] Module [\{memory\_cap\}\% --- \{memory\_char\}/\{upper\_memory\} chars]: \\
\{memory\_info\} \\
 \\
(6) Current [RETRIEVAL MEMORY] Module [\{missing\_cap\}\% --- \{missing\_char\}/\{upper\_missing\} chars]: \\
\{missing\_info\} \\
 \\
**INSTRUCTIONS:** \\
Reflect: Comprehensively compare the forecasting process and predicted unit prices point-by-point against the ground-truth trajectory, then update each module. Keep all entries generalizable and transferable across \{name\} parts and periods --- not SKU- or month-specific: \\
1. \texttt{<SELECTION\_MEMORY>} (event SELECTION) --- general criteria for *which event types* most move \{name\} prices and which are noise. \\
2. \texttt{<RETRIEVAL\_MEMORY>} (event SUPPLEMENTATION) --- broad categories / recurring blind spots of forecast-moving events absent from the pool, to seek next time. \\
 \\
Add ONLY *new* insights, lessons, or experiences gained this round. Check it against the existing module contents above and OMIT experiences already covered and use ``\S'' as the separator between each piece of insight in tags. \\
 \\
**OUTPUT FORMAT:** \\
{}[Comprehensive Comparison \& Analysis] \\
\texttt{<SELECTION\_MEMORY>} ... \texttt{</SELECTION\_MEMORY>} \\
\texttt{<RETRIEVAL\_MEMORY>} ... \texttt{</RETRIEVAL\_MEMORY>}
\end{AIbox}
\normalsize
\captionof{figure}{The SEER memory learning prompt template (Part A) for price forecasting tasks, which updates the selection memory [SELECTION MEMORY] and the retrieval memory [RETRIEVAL MEMORY]. Placeholders adapt the template to a specific domain's contextual cues. For example, in the computer memory domain, \{name\} corresponds to ``RAM'', \{reflect\_dynamics\} to ``AI-server CapEx, opportunity-cost parity'', \{missing\_signals\} to ``raw fab utilization, hidden capacity reallocations'', and \{sector\} to ``Semiconductor Memory / RAM Products''. The remaining placeholders represent the inputs detailed in Section~\ref{sec:method}. For non-price forecasting tasks, we follow the same template and learning structure, with adaptations to specific descriptions (e.g., modifying ``ground-truth prices'').}
\label{fig:seer-learn-mm}
\endgroup

\clearpage

\begingroup
\tcbset{breakable,enhanced jigsaw}
\scriptsize
\setlength{\parindent}{0pt}

\begin{AIbox}{{SEER Memory Learning (Part B) --- System Prompt: [CAUSAL KNOWLEDGE]}}
**TASK:** Optimize the [CAUSAL KNOWLEDGE] module by comparing the LLM's monthly **\{name\} unit-price** forecast against the realized ground-truth prices. \\
([CAUSAL KNOWLEDGE] = lessons learned --- durable sector-level rules and reasoning frameworks distilled from this run's hits and misses.) \\
* **Reflect:** Compare the predicted monthly unit prices and the LLM's reasoning trace point-by-point against the ground-truth price trajectory. Identify which events / mechanisms drove correct vs. incorrect movements, and what magnitude or direction errors say about overlooked structural dynamics (e.g., \{reflect\_dynamics\}). \\
* **Context:** Everything provided below for this reflection --- the forecasting conditions, the prediction results (forecasting process + ground-truth prices), and the current state of [CAUSAL KNOWLEDGE]. \\
* **Update [CAUSAL KNOWLEDGE] (lessons learned):** Turns past correct/incorrect predictions into better future accuracy. Record high-level rules of thumb, strategic takeaways, and analytical frameworks distilled from this run's hits and misses --- durable across \{name\} parts and periods.
\end{AIbox}
\begin{AIbox}{{SEER Memory Learning (Part B) --- User Prompt: [CAUSAL KNOWLEDGE]}}
(1) Prediction Conditions: \\
**Target Part:** \{company\_name\} \\
**Cut-off Month:** \{fore\_cut\_off\_date\} \\
**Prediction Horizon:** Next \{prediction\_horizon\} months unit price \\
 \\
**A. Historical Monthly Unit Prices (From \{start\_trading\_date\} to \{fore\_cut\_off\_date\})** \\
\{his\_prices\} \\
 \\
**B. Events Selected** \\
\{event\_list\_selected\} \\
 \\
(2) Prediction Process \& Label: \\
\{llm\_forecasting\_resp\} \\
 \\
**Ground-Truth Monthly Unit Prices:** \{label\} \\
 \\
(4) [SECTOR] Information: \\
\{sector\} \\
 \\
(5) Current [CAUSAL KNOWLEDGE] Module [\{knowledge\_cap\}\% --- \{knowledge\_char\}/\{upper\_know\} chars]: \\
\{knowledge\_info\} \\
 \\
**INSTRUCTIONS:** \\
Reflect: Comprehensively compare the forecasting process and predicted unit prices point-by-point against the ground-truth trajectory, then update the module. Keep all entries generalizable and transferable across \{name\} parts and periods --- not SKU- or month-specific: \\
1. \texttt{<CAUSAL\_KNOWLEDGE>} (lessons learned) --- high-level rules of thumb, strategic takeaways, and analytical frameworks distilled from this run's hits and misses. \\
 \\
Add ONLY *new* insights, lessons, or experiences gained this round. Check it against the existing module contents above and OMIT experiences already covered and use ``\S'' as the separator between each piece of insight in tags. \\
 \\
**OUTPUT FORMAT:** \\
{}[Comprehensive Comparison \& Analysis] \\
\texttt{<CAUSAL\_KNOWLEDGE>} ... \texttt{</CAUSAL\_KNOWLEDGE>}
\end{AIbox}
\normalsize
\captionof{figure}{The prompt template of SEER \emph{KNOWLEDGE} learning, Part B, which updates the causal forecasting knowledge [CAUSAL KNOWLEDGE]. The placeholders share the identical domain-specific adaptations as detailed in Figure~\ref{fig:seer-learn-mm}.}
\label{fig:seer-learn-know}
\endgroup

\clearpage 

\begingroup
\tcbset{breakable,enhanced jigsaw}
\small
\setlength{\parindent}{0pt}

\begin{AIbox}{{SEER Memory Condense --- [SELECTION MEMORY]}}
\textit{System prompt:} \\
Your task is to optimize a ``Memory List'' containing historical market drivers and predictive factors for the sector: \\
\{sector\} \\
 \\
Guidelines for condensing the memory: \\
1. Prune and Refine problematic rules: Remove highly specific rules or those explicitly flagged as flawed, misleading, or outdated, or modify them based on refinement and reflection feedback. \\
2. Merge the redundant: Identify overlapping concepts or highly similar factors and synthesize them into a single, comprehensive rule. \\
3. Preserve the most actionable, sector-wide rules: Do not oversimplify the useful rules that are retained; maintain their necessary depth and operational nuance. \\
 \\
First, perform a comprehensive analysis of your knowledge condensation. Then, place the FINAL COMPLETE updated knowledge list in order from 1 to N inside sequentially numbered tags. \\
 \\
**OUTPUT FORMAT:** \\
{}[Comprehensive Reasoning \& Analysis] \\
 \\
\texttt{<SELECTION\_MEMORY1>} ... \texttt{</SELECTION\_MEMORY1>} \\
\texttt{<SELECTION\_MEMORY2>} ... \texttt{</SELECTION\_MEMORY2>} \\
... \\
\texttt{<SELECTION\_MEMORYN>} ... \texttt{</SELECTION\_MEMORYN>}

\tcbline

\textit{User prompt:} \\
Here is the current memory list: \\
\{memory\_str\}
\end{AIbox}
\begin{AIbox}{{SEER Memory Condense --- [RETRIEVAL MEMORY]}}
\textit{System prompt:} \\
Your task is to optimize a ``Missing Signals List'' containing suggestions for uncaptured predictive factors and information gaps that could help provide supplementary events to assist forecasting for the sector: \\
\{sector\} \\
 \\
Guidelines for condensing the missing signals: \\
1. Elevate to Sector-Level: Generalize highly specific suggestions into broader, sector-wide signals. Remove narrow metrics that lack structural relevance to the overall sector. \\
2. Prune or Refine problematic insights: Remove highly specific insights or those explicitly flagged as flawed, misleading, or outdated, or modify them based on refinement and reflection feedback. \\
3. Merge the redundant: Identify overlapping concepts or highly similar missing factors and synthesize them into a single, comprehensive signal. \\
4. Preserve the most actionable, sector-wide insights: Do not oversimplify the useful insights that are retained; maintain their necessary depth and operational nuance. \\
 \\
First, perform a comprehensive analysis of your knowledge condensation. Then, place the FINAL COMPLETE updated knowledge list in order from 1 to N inside sequentially numbered tags. \\
 \\
**OUTPUT FORMAT:** \\
{}[Comprehensive Reasoning \& Analysis] \\
 \\
\texttt{<RETRIEVAL\_MEMORY1>} ... \texttt{</RETRIEVAL\_MEMORY1>} \\
\texttt{<RETRIEVAL\_MEMORY2>} ... \texttt{</RETRIEVAL\_MEMORY2>} \\
... \\
\texttt{<RETRIEVAL\_MEMORYN>} ... \texttt{</RETRIEVAL\_MEMORYN>}

\tcbline

\textit{User prompt:} \\
Here is the current list of missing signals: \\
\{memory\_str\}
\end{AIbox}
\begin{AIbox}{{SEER Memory Condense --- [CAUSAL KNOWLEDGE]}}
\textit{System prompt:} \\
Your task is to produce the FINAL, COMPLETE, updated ``Knowledge List'' of durable sector-level forecasting rules for the sector: \\
\{sector\} \\
 \\
Guidelines for condensing the memory: \\
1. Prune and Refine problematic rules: Remove highly specific rules or those explicitly flagged as flawed, misleading, or outdated, or modify them based on refinement and reflection feedback. \\
2. Merge the redundant: Identify overlapping concepts or highly similar factors and synthesize them into a single, comprehensive rule. \\
3. Preserve the most actionable, sector-wide rules: Do not oversimplify the useful rules that are retained; maintain their necessary depth and operational nuance. \\
 \\
First, perform a comprehensive analysis of your knowledge condensation. Then, place the FINAL COMPLETE updated knowledge list in order from 1 to N inside sequentially numbered tags. \\
 \\
**OUTPUT FORMAT:** \\
{}[Comprehensive Reasoning \& Analysis] \\
 \\
\texttt{<CAUSAL\_KNOWLEDGE1>} ... \texttt{</CAUSAL\_KNOWLEDGE1>} \\
\texttt{<CAUSAL\_KNOWLEDGE2>} ... \texttt{</CAUSAL\_KNOWLEDGE2>} \\
... \\
\texttt{<CAUSAL\_KNOWLEDGEN>} ... \texttt{</CAUSAL\_KNOWLEDGEN>}

\tcbline

\textit{User prompt:} \\
**Learned Knowledge List (numbered, in order):** \\
\{existing\_knowledge\}
\end{AIbox}
\normalsize

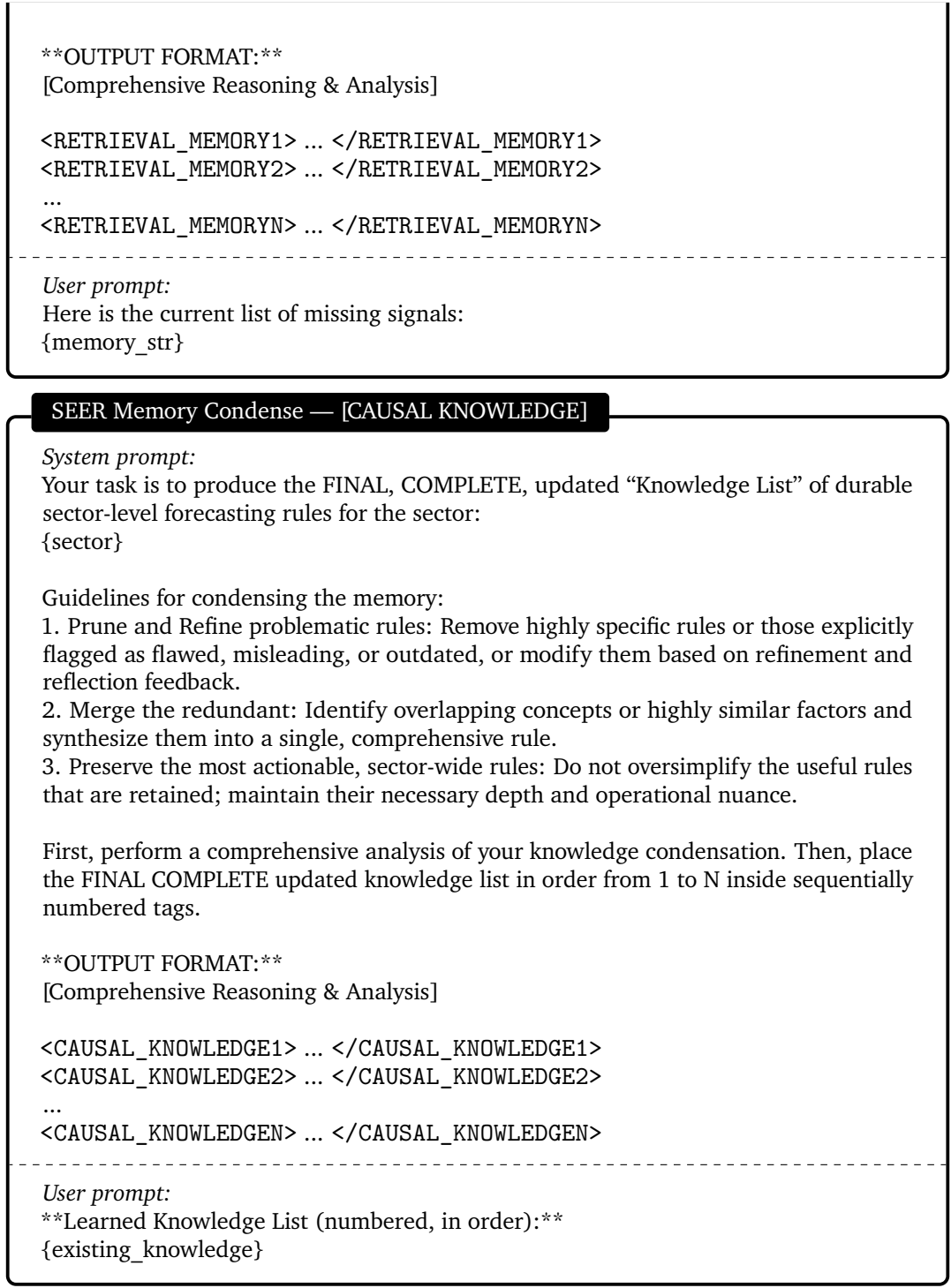
\captionof{figure}{Prompt templates of SEER memory condense, one box per module, with the user prompt below the system prompt. Within the user prompt, the previously accumulated contextual memories are enumerated sequentially from 1 to $N$. After a learning pass, a module is condensed once its character count exceeds 80\% of its ceiling, i.e.\ of the value shown to the learner as \{upper\_memory\}, \{upper\_missing\} or \{upper\_know\} in Figures~\ref{fig:seer-learn-mm} and \ref{fig:seer-learn-know}: 12,000 characters for [SELECTION MEMORY], 14,000 for [RETRIEVAL MEMORY] and 22,000 for [CAUSAL KNOWLEDGE]).}
\label{fig:seer-condense}
\endgroup

\clearpage

\begingroup
\tcbset{breakable,enhanced jigsaw}
\small
\setlength{\parindent}{0pt}

\begin{AIbox}{{SEER Event Search (Initial) --- System Prompt}}
You are an event analysis assistant. Your task is to execute a comprehensive search to identify any material events that could influence the future trend of a specific target. \\
 \\
**Analysis \& Classification Rules:** \\
* **Causal Depth:** Explain the *why* via second-order effects, not just surface headlines. Capture all positive and negative factors. \\
* **Tagging Constraints:** \\
\hspace*{2.0em}* **Sentiment:** Strictly limit to \texttt{\textasciigrave{}Positive\textasciigrave{}} or \texttt{\textasciigrave{}Negative\textasciigrave{}}. \\
\hspace*{2.0em}* **Impact Type:** - If the event directly involves the target -\textgreater{} tag as \texttt{\textasciigrave{}Direct\textasciigrave{}}. \\
\hspace*{4.0em}- If the event involves related entities or the broader environment -\textgreater{} tag as \texttt{\textasciigrave{}Indirect\textasciigrave{}}.
\end{AIbox}
\begin{AIbox}{{SEER Event Search (Initial) --- User Prompt (Round 1; dashed block for Round $\geq 2$)}}
\tikz[baseline=-0.6ex]\draw[dashed,gray!80] (0,0) -- (\linewidth,0); \\
\textcolor{gray}{\textit{Round $\geq$ 2 only: the events found in the previous round(s) are inserted here so that they are not repeated.}} \\
The following events have been identified: \\
\{events\_lists\} \\
\tikz[baseline=-0.6ex]\draw[dashed,gray!80] (0,0) -- (\linewidth,0); \\
 \\
Leverage your knowledge about ``\{target\_name\}'', perform an exhaustive search for all events influencing its future trend on the specific date ``\{target\_date\}''. \\
 \\
**Primary Target:** Identify all impactful events first released on '\{target\_date\}'. \\
 \\
**Output JSON Format:** \\
\texttt{\textasciigrave{}\textasciigrave{}\textasciigrave{}json} \\
{}[ \\
\hspace*{1.0em}\{ \\
\hspace*{2.0em}``date'': ``YYYY-MM-DD'', \\
\hspace*{2.0em}``description'': ``A specific, fact-based summary explicitly naming the exact actor and their concrete action, quote, or data point.'', \\
\hspace*{2.0em}``causality'': ``Breifly explaining the transmission mechanism: why and how this event moved the stock price.'', \\
\hspace*{2.0em}``sentiment'': ``Positive'' \textbar{} ``Negative'', \\
\hspace*{2.0em}``impact\_type'': ``Direct'' \textbar{} ``Indirect'' \\
\hspace*{1.0em}\}, \\
\hspace*{1.0em}... (List all identified factors) \\
] \\
\texttt{\textasciigrave{}\textasciigrave{}\textasciigrave{}}
\end{AIbox}
\normalsize
\captionof{figure}{Prompt template for the SEER initial event search in the stock price forecasting domain. Given the next-day horizon for stock prediction, the search is conducted daily: \{target\_date\} restricts results to events published on that specific day. For other tasks (e.g., memory prices or electricity demand), the search spans a 7-day window with explicit start and end dates, while maintaining the same overall template structure. The placeholder \{target\_name\} denotes the company name resolved from its ticker symbol (e.g., ``NVIDIA Corporation'' for NVDA). The search is multi-round: from round~2 onwards, the block between the dashed lines is appended to the user prompt, listing prior events (\{events\_lists\}) to avoid duplicates. Note that the forecasting inputs include only the event descriptions. Furthermore, only events that achieve consensus via the multi-agent fact-checking module in LEAF~\cite{tan2026leaf}, along with their verified dates, are provided. A concrete example is provided in \autoref{fig:64gb-ddr5-prompt}.}
\label{fig:seer-search-init}
\endgroup

\clearpage

\begingroup
\tcbset{breakable,enhanced jigsaw}
\scriptsize
\setlength{\parindent}{0pt}

\begin{AIbox}{{SEER Event Further Search--- System Prompt}}
Your objective is to review a preliminary stock events and fill in critical information gaps based on ``Missing Info Clues''. \\
 \\
You will be provided with an existing list of events and a set of clues pointing to overlooked factors. Your mandate is to uncover, search, and identify the missing events without duplicating existing entries. \\
 \\
**Analysis \& Classification Rules:** \\
* **Causal Depth:** Explain the *why* via second-order effects, not just surface headlines. \\
* **Tagging Constraints:** \\
\hspace*{2.0em}* **Sentiment:** Strictly limit to \texttt{\textasciigrave{}Positive\textasciigrave{}} or \texttt{\textasciigrave{}Negative\textasciigrave{}}. \\
\hspace*{2.0em}* **Impact Type:** - If the event directly involves the target company -\textgreater{} tag as \texttt{\textasciigrave{}Direct\textasciigrave{}}. \\
\hspace*{4.0em}- If the event involves related entities, sector peers, or macro environment -\textgreater{} tag as \texttt{\textasciigrave{}Indirect\textasciigrave{}}.
\end{AIbox}
\begin{AIbox}{{SEER Event Further Search --- User Prompt}}
Target Entity/Company: ``\{target\_name\}'' \\
Target Date: ``\{target\_date\}'' \\
 \\
**Existing Intelligence (Already Captured):** \\
The following events have already been identified before \{target\_date\}. \\
\{events\_lists\} \\
 \\
**New Intelligence (Missing Info Clues):** \\
\{missing\_info\} \\
 \\
**Task:** \\
Leverage your knowledge and deeply analyze the ``Missing Info'' clues. Perform a targeted search to identify the exact material events, data points, or catalyst shifts that were missed for ``\{target\_name\}'' first released on ``\{target\_date\}''. \\
 \\
**Content Requirements:** \\
For the ``description'' field, you must construct a specific summary of the event: \\
* The Actor: The specific entity (e.g., AMD or Google), person (e.g., CEO name), or institution (e.g., ``The Fed'') involved. \\
* The Action/Data: The specific event or number (e.g., ``resigned,'' ``raised rates by 25bps,'' ``missed revenue by 5\%''). \\
* Avoid vague phrases. \\
 \\
**Output the final results strictly in the following JSON format:** \\
\texttt{\textasciigrave{}\textasciigrave{}\textasciigrave{}json} \\
{}[ \\
\hspace*{1.0em}\{ \\
\hspace*{2.0em}``date'': ``YYYY-MM-DD'', \\
\hspace*{2.0em}``description'': ``A specific, fact-based summary explicitly naming the exact actor and their concrete action, quote, or data point.'', \\
\hspace*{2.0em}``causality'': ``Breifly explaining the transmission mechanism: why and how this event moved the stock price.'', \\
\hspace*{2.0em}``sentiment'': ``Positive'' \textbar{} ``Negative'', \\
\hspace*{2.0em}``impact\_type'': ``Direct'' \textbar{} ``Indirect'' \\
\hspace*{1.0em}\}, \\
\hspace*{1.0em}... (List all identified factors) \\
] \\
\texttt{\textasciigrave{}\textasciigrave{}\textasciigrave{}}
\end{AIbox}
\normalsize
\captionof{figure}{As detailed in Section~\ref{sec:method-env}, the retrieval memory $\mathcal{M}_t^{\text{ret}}$ guides the expanded event search. The prompt template, placeholders and search window configurations are similar as \autoref{fig:seer-search-init} ``initial search'', but incorporates the learned [RETRIEVAL MEMORY] into the \{missing\_info\} placeholder. Expanded event searches across other domains follow a similar template structure. For specialized domains, a general domain description can be provided to focus the search better on relevant events. }
\label{fig:seer-search-missing}
\endgroup


\clearpage

\begingroup
\tcbset{breakable,enhanced jigsaw}
\tiny
\setlength{\parindent}{0pt}
\begin{AIbox}{{SEER Learned Knowledge [CAUSAL KNOWLEDGE] --- Memory unit-price forecasting}}
1. **Regime Classification \& Foundational Drivers** \\
Before forecasting any SKU, explicitly classify the regime (Squeeze vs. Glut) using a 4-point checklist to avoid catastrophic sign errors: \\
1. Is supply expanding (rising utilization/wafer output) or contracting? \\
2. Are downstream buyers accepting quoted price hikes or resisting/threatening non-purchase? \\
3. Is supplier guidance narrowing (loosening) or widening (tightening)? \\
4. Is the specific SKU trailing-edge (glut-prone) or leading-edge? \\
*Verdict:* If supply is contracting and buyers are accepting hikes (or ``sold out'' announcements occur), it is an UPSIDE squeeze. If supply is expanding and buyers resist, it is a DOWNSIDE cycle top. \\
*Corollaries:* \\
- CEO/manufacturer forward pricing language (``prices rising throughout the year'') strongly overrides single-quarter analyst guidance. \\
- In a structural supply-deficit, price is set at the margin by supply scarcity; weak conventional demand does NOT cap prices. \\
- Spot market weakness (e.g., spot DDR4 drifting down) is transient retail noise; ignore it completely when forecasting CONTRACT SKUs. \\
\S \\
2. **Execution Hierarchy \& Math Discipline** \\
Forecast error is dominated by a strict precision hierarchy: (1) Correct regime sign, (2) Correct base selection, (3) Correct step timing (Q-boundary placement), and (4) Multiplier magnitude. \\
- *Target Math:* Compute the target strictly as: \texttt{\textasciigrave{}Target = Correct\_Base $\times$ Precedent\_Multiplier\textasciigrave{}}. \\
- *Rounding Discipline:* Carry one extra significant figure through the multiplication (e.g., Base $\times$ 1.780) before rounding the final target to two decimals. This shaves residual basis points and ensures cent-exact accuracy. \\
- *Error Propagation:* Base errors propagate multiplicatively. Base-inflation compounds magnitude misses: overshooting the pre-boundary base while simultaneously under-applying the boundary multiplier produces a catastrophic ``double miss.'' Always anchor the structural leg to the correct SETTLED base. \\
- *Commitment:* Once the base, timing, and sign are locked, apply the SKU-specific realized multiplier with complete commitment. Do not artificially hedge, second-guess, or fine-tune to chase sub-0.3\% precision. \\
\S \\
3. **Base Calibration \& Handling Guidance** \\
- *Settled Base:* If historical prints show the current quarter settled, carry that exact value forward flat to two decimals. Do NOT inflate it mid-quarter. \\
- *Synthesized Base from Guidance:* \\
\hspace*{1.0em}- *Modest Guidance (e.g., 3-8\%):* Settlements tend to clear near the arithmetic MIDPOINT. \\
\hspace*{1.0em}- *Demand Caveat:* If suppliers/analysts explicitly note demand ``did not materialize as anticipated'' or guidance was ``narrowed,'' the settlement clears in the LOWER HALF of the range (e.g., \textasciitilde{}4\%), NOT the absolute minimum (\textasciitilde{}3\%). Only push to the true floor when demand language is unambiguously weak AND supply is loosening. \\
- *Aggregate vs. SKU:* Do NOT convert an industry-aggregate contract-price \% (e.g., TrendForce ``13-18\% Q2 DRAM increase'') into a per-SKU price step. Aggregate figures blend many parts; the SKU's OWN historical prints already encode its settled move. Fabricating an intra-quarter lift to reconcile a SKU print with an aggregate headline is a critical failure mode. \\
\S \\
4. **Structural Multipliers \& Wafer Cannibalization** \\
Wafer reallocation to premium products creates violent, discontinuous price spikes. HBM consumes leading-edge wafers at a 3:1 bit ratio, starving both leading-edge and trailing-edge SKUs. Trailing-edge starvation often produces equal or greater price explosions due to severe capacity cuts and inelastic demand. \\
*First Full Structural Leg Multipliers:* Match the target SKU on node, capacity, and form-factor. \\
- Leading-edge DDR5 Server RDIMM: \textasciitilde{}1.78x (Realized exactly 1.780x. Do not use the older \textasciitilde{}1.807x ladder value for this SKU). \\
- Trailing-edge DDR4 32GB Module: \textasciitilde{}1.845x \\
- Scarcest Trailing-edge DDR4 64GB RDIMM: \textasciitilde{}1.880x \\
*Shock Amplification:* When a structural step is driven by a one-time physical shock (earthquake, fab outage) combined with a \textgreater{}20-25\% manufacturer hike notice, the realized settlement (e.g., \textgreater{}78\% jump) far EXCEEDS the nominal hike-notice \%. The hike notice acts as a floor, and the physical shock amplifies it. \\
\S \\
5. **Shape, Timing, and Plateau Discipline (Step-Then-Plateau)** \\
Contract-priced RAM SKUs move in discontinuous steps. The default shape is a flat base, ONE large discrete step, and then perfectly FLAT again. \\
- *Current Quarter Hold:* Hold the CURRENT quarter perfectly flat at the settled print. Even if your reasoning trace explicitly identifies an accelerating deficit ramping into the next quarter, acceleration expresses itself at the NEXT Q-boundary, not by retroactively inflating the already-settled current quarter. \\
- *Timing the Step:* Place the full structural step exactly at the FIRST month of the NEXT quarter (Q-boundary). The ABSENCE of published next-quarter analyst guidance is NEVER a valid reason to hold flat into an identified accelerating deficit. Structural squeezes reprice on physical scarcity, not on whether analysts have quantified a number. \\
- *Plateau Discipline:* Once a contract reprices, it holds perfectly flat. Never sprinkle small monthly gains (``marginal firmness'') onto a plateau, never apply a reflexive decay tick, and ignore sub-0.2\% micro-wobble within a settled quarter. One-time discrete physical shocks (earthquakes) ALWAYS produce a single step followed immediately by a flat plateau. \\
\S \\
6. **Multi-Leg Squeezes \& Leg-Counting Discipline** \\
Structural squeezes can arrive as multiple sequential legs, but you must strictly categorize prior steps to assign the correct multiplier. \\
- *Leg-Counting:* Modest single-digit intra-quarter steps (+3-8\%) do NOT count as ``structural legs.'' If the prior quarter merely settled at a mild step, the next Q-boundary jump is a FIRST full structural leg and clears at the \textasciitilde{}1.78x - 1.88x floor (See Knowledge 4). \\
- *Second Leg Multiplier:* If the base already contains a violent structural spike (\textasciitilde{}1.8x), the NEXT structural leg at a later Q-boundary clears at the upper edge of the secondary band: \textasciitilde{}1.43x. \\
- *Within-Quarter Staircase (Rare Exception):* Default shape is always Step-Then-Plateau. A high accelerant count of long-horizon narrative items (new fab CapEx, ``sold out through 2025,'' GPU roadmaps) supports the FIRST leg and the plateau level---they do NOT generate fresh monthly steps. ONLY real-time worsening of a deficit within the quarter (rising HBM capacity share prints, sequential real-time hike notices) justifies a second within-quarter leg (\textasciitilde{}1.18x month-over-month). When uncertain, strongly prefer the plateau; over-forecasting multiple legs based on narrative backdrop is a primary failure mode.
\end{AIbox}
\normalsize
\captionof{figure}{The final causal knowledge [CAUSAL KNOWLEDGE] learned by SEER (Claude Opus 4.8) in the memory domain, printed exactly as it is handed to the forecasting prompt (such as ~\autoref{fig:64gb-ddr5-prompt}): 6 numbered rules separated by \S.}
\label{fig:seer-knowledge-ram}
\endgroup

\clearpage

\begingroup
\tcbset{breakable,enhanced jigsaw}
\tiny
\setlength{\parindent}{0pt}
\begin{AIbox}{{SEER Learned Knowledge [CAUSAL KNOWLEDGE] --- cross-stock batch forecasting (14 of 66 rules)}}
1. IF a stock enters an earnings/guidance catalyst after a cluster of pre-event price-target hikes, ``top pick'' tags, and a price series already printing fresh highs into the report THEN reduce conviction on a bullish beat-and-raise (or flip toward fade) because the positive outcome is likely priced-in and invites a sell-the-news / crowding unwind (scope: market-wide, momentum/late-cycle regime, earnings + analyst-action) \\
\S \\
2. IF a stock reports a beat-and-raise while its recent event pool is dominated by bearish signals (price-target cuts, fund liquidations, prominent bear commentary) THEN raise conviction on continuation upside because the surprise resets a skeptical crowd's positioning and forces short-covering/re-rating (scope: market-wide, any regime, earnings + analyst-action) \\
\S \\
3. IF a stock's bullish thesis rests on a commodity/macro price shock that has ALREADY moved the underlying and the stock has closed at a window high on the same session the catalyst broke THEN reduce conviction or fade the next-day continuation, because the initial shock is reflected in the current close and the follow-through tends to be reversion/exhaustion rather than fresh repricing (scope: energy/commodity-linked sectors, any regime, macro/geopolitical price-shock) \\
\S \\
4. IF two or more of the batch's top picks derive their directional thesis from the same underlying driver (same commodity, same macro print, same sector catalyst) THEN treat them as a single correlated bet and cap combined conviction, because the apparent multi-pick confidence is illusory and one thesis reversal takes down all correlated picks simultaneously (scope: market-wide, any regime, sector/macro catalyst) \\
\S \\
5. IF a pick's bullish thesis rests on a single-session outsized gap driven by a company-specific catalyst that is already fully reflected in the current close (a new customer/order win, product news, or a multi-day post-earnings recovery that has round-tripped back to fresh highs) THEN reduce conviction or fade next-day continuation, because the catalyst is spent and the parabolic session invites profit-taking/mean-reversion rather than fresh repricing (scope: market-wide, momentum/late-cycle regime, product-order or earnings-recovery catalyst) \\
\S \\
6. IF the same session that produced a pick's bullish catalyst also contains an explicit profit-taking or insider/analyst sell-to-lock-gains signal (or prominent valuation-overextension commentary) in its event pool THEN treat that as a direct exhaustion tell and downgrade the pick toward Neutral/Down, because informed participants are already unwinding into the strength and the marginal buyer is thinning (scope: market-wide, any regime, catalyst-driven breakout) \\
\S \\
7. IF a mega-cap raises capital-expenditure/spending guidance in the same report as a revenue beat, THEN predict Down if the report lacks concurrent proof of accelerating monetization (segment backlog growth, cloud/unit acceleration) and instead pairs the spend with margin-drag or loss-making segment commentary, but lean Up if the same report shows the spend is already converting into accelerating high-margin revenue, because the market prices rising CapEx as value-destructive absent visible ROI and value-accretive when acceleration confirms the return (scope: technology/semiconductors, momentum/late-cycle regime, earnings + guidance) \\
\S \\
8. IF a stock enters earnings already at fresh highs with a pre-event price-target-hike cluster (the Rule 1 fade setup) BUT then delivers a beat whose surprise is broad-based and accompanied by explicit acceleration in a key growth segment, THEN override the priced-in fade and lean toward continuation Up, because a large positive surprise magnitude combined with fresh acceleration forces upward re-rating faster than crowded positioning can unwind (scope: market-wide, momentum/late-cycle regime, earnings + analyst-action) \\
\S \\
\textcolor{gray}{\textit{\ldots\ (rules 9--63 omitted) \ldots}} \\
\S \\
64. IF a pick's structural bull catalysts (sold-out capacity, upgrade cluster) are dated a session BEFORE the cutoff and the cutoff session itself is an outsized parabolic single-bar advance that absorbs them THEN do NOT apply the sold-out/cutoff-dated continuation override; lean Neutral, because the durable re-rate has been front-loaded into one exhaustion bar and the marginal buyer is thinning rather than arriving next session (scope: technology/semiconductors, momentum/late-cycle regime, earnings + structural-demand catalyst) \\
\S \\
65. IF a pick's bullish thesis rests on a fresh cutoff-dated analyst upgrade/PT-hike while the SAME company's most recent earnings print was a genuine miss-and-cut (sales miss and/or LOWERED forward guidance) rather than a beat-and-raise, THEN do NOT apply the post-print continuation override (Rule 43); lean Neutral, because a lone derivative opinion catalyst cannot manufacture a fresh forward-earnings fact from deteriorating fundamentals, and layering it on a tape that already popped stalls once the upgrade headline is absorbed (scope: market-wide, any regime, earnings-miss + analyst-action) \\
\S \\
67. IF a beaten-down stock in a multi-session decline shows a fresh cutoff-dated DISCRETIONARY open-market insider PURCHASE (an officer/director committing personal capital at the prevailing depressed price, not a scheduled 10b5-1 execution) THEN lean continuation Up with raised magnitude, because a discretionary buy at current levels is a direct forward-value signal from an informed party that resets a skeptical, oversold crowd's positioning and forces re-rating/short-covering --- the bullish mirror of the insider-sale exhaustion tell (scope: market-wide, any regime, catalyst + insider-transaction) \\
\S \\
68. IF a beaten-down stock's cutoff-dated event pool STACKS independent bullish drivers --- a discretionary insider buy, a validated product launch with strong third-party reviews, AND a fresh analyst upgrade cluster --- landing together on an oversold tape, THEN raise conviction toward strong continuation Up, because these are non-overlapping unpriced signals (positioning reset + forward-earnings-ceiling repricing + sell-side catch-up) that compound across sessions rather than a single spent catalyst that reverts (scope: technology/semiconductors, any regime, insider-transaction + product-launch + analyst-action) \\
\S \\
69. IF a mega-cap's cutoff-session earnings crack the shared demand narrative that also underpins a pick's bull thesis (e.g., a hyperscaler reporting cloud deceleration plus capex-return doubts, while the pick is a downstream capex-beneficiary such as semicap or memory) THEN downgrade the pick toward Neutral/Down rather than treating its sub-sector as insulated, because the market re-prices the entire demand-chain of a broken narrative in one session and downstream beneficiaries fall with the source even when their own print was a beat (scope: technology/semiconductors, risk-off/rotation regime, peer-spillover + narrative-repricing catalyst) \\
\S \\
70. IF two of a batch's bull picks both derive their thesis from the same upstream capex/demand driver AND both enter the cutoff on parabolic tapes at window highs, THEN treat them as one correlated late-stage bet and cap combined conviction toward Neutral even with fresh post-print upgrade clusters, because a single narrative reversal simultaneously repricing the shared driver hits the extended tapes hardest and the upgrade catch-up is overwhelmed by the correlated outflow (scope: technology/semiconductors, momentum/late-cycle regime, earnings + analyst-action correlated-basket catalyst)
\end{AIbox}
\normalsize
\captionof{figure}{The final causal knowledge [CAUSAL KNOWLEDGE] learned by SEER (Claude Opus 4.8) for cross-stock batch forecasting: 70 conditional IF--THEN rules in the same format, numbered 1--70 as stored and as the forecasting prompt cites them (Figure~\ref{fig:stockbatch-prompt}); Only the first 8 and the last 6 rules are shown; rules 9--63 are omitted for space.}
\label{fig:seer-knowledge-stockbatch}
\endgroup

\clearpage

\begingroup
\tcbset{breakable,enhanced jigsaw}
\small
\setlength{\parindent}{0pt}
\begin{AIbox}{{SEER Learned Electricity Knowledge [CAUSAL KNOWLEDGE]}}
1. IF the trailing days at the edge of the forecast window were suppressed by a Public Safety Power Shutoff (PSPS) or other forced-outage event, THEN anchor the forecast MAX and MIN above those suppressed trailing values, back toward the pre-outage observed level, because outage-depressed readings understate underlying demand and the load rebounds to its pre-event level once power is restored. (scope: San Francisco) \\
\S \\
2. IF a severe heat event peaked prior to the forecast window and the trailing-week daily MAX is actively easing, THEN step the forecast MAX and MIN down steeply toward recent pre-heatwave trailing lows rather than holding them elevated, because releasing accumulated cooling load causes underlying demand to decay faster than standard temperature regressions predict. (scope: New York) \\
\S \\
3. IF a winter storm clears early in the forecast window but is followed by an intrusion of lingering Arctic air (highs held low, overnight lows depressed), THEN hold the post-storm MAX and MIN at or above their pre-storm observed levels rather than stepping them down, because the lingering cold air mass sustains structural heating demand even after precipitation ends. (scope: New York, San Francisco) \\
\S \\
4. IF a severe weather event (heat wave or deep winter cold) spans a weekend or holiday, THEN hold the weekend MAX and MIN near the surrounding weekday peak/trough rather than applying the standard calendar markdown, because intense residential HVAC demand overrides the typical commercial schedule drop. (scope: New York, San Francisco) \\
\S \\
5. IF the forecast window follows a cold front or rainy spell and the NWS forecasts a sustained multi-day sunny warming trend building through the window, THEN step the forecast MAX progressively upward day-over-day above the recent trailing peaks rather than holding flat, because early-season cooling load accumulates across consecutive warming days and lifts daytime peaks faster than a single-day temperature regression suggests. (scope: San Francisco) \\
\S \\
6. IF a heat event peaked just before the forecast window and the trailing daily MAX is easing, THEN hold the forecast overnight MIN up near the recent elevated trailing MIN rather than stepping it down proportionally with the MAX, because a warm residual air mass keeps the overnight minimum load from decaying even after daytime cooling peaks have receded. (scope: San Francisco) \\
\S \\
7. IF the trailing edge of the forecast window is a weekend (or holiday) showing depressed MAX and MIN, and the window opens with a run of weekdays, THEN anchor the weekday MAX and MIN to the most recent WEEKDAY levels rather than lifting only modestly above the weekend trough, because the weekend markdown understates the underlying weekday demand base and using it as the anchor systematically suppresses the whole weekday block. (scope: San Francisco, summer, cool coastal marine climate)
\end{AIbox}
\normalsize
\captionof{figure}{The final causal knowledge [CAUSAL KNOWLEDGE] learned by SEER (Claude Opus 4.8) for city-level daily electricity-load forecasting: 7 rules, the state after the sequential learning pass over the three cities (Boston, New York, San Francisco; last cut-off 2025-07-27).}
\label{fig:seer-knowledge-elec}
\endgroup

\clearpage

\ifcsname AIbox\endcsname\else
  \newtcolorbox{AIbox}[2][]{colback=gray!5!white,colframe=black!60!white,fonttitle=\bfseries,
    title=#2,boxrule=0.5pt,arc=1.5pt,left=3pt,right=3pt,top=2pt,bottom=2pt,#1}
\fi
\begingroup
\tcbset{breakable,enhanced jigsaw}
\tiny
\setlength{\parindent}{0pt}
\begin{AIbox}{{System Prompt For Memory Price Forecasting}}
**Task:** \\
You will be provided with a RAM product's recent monthly **unit prices** and a list of news events (or N/A) affecting that part. Your task is to forecast the **exact monthly unit prices** (in USD) for the next 9 consecutive months following the cut-off month. \\
 \\
**Notes:** \\
* Prices are observed monthly; each value represents the unit price for that month. \\
* The prediction targets are the next 9 specific months listed below; you must emit one price per target month, in chronological order. \\
* Prediction Knowledge is the historical forecasting experience; please refer to it when making predictions. \\
\end{AIbox}
\begin{AIbox}{{User Prompt For 128GB DDR5 at Cut-off 2025-06 (18 of 66 events shown, 6 knowledge rules)}}
**Target Part:** 128GB-DDR5 \\
**Cut-off Month:** 2025-06 \\
**Prediction Horizon:** Next 9 months \\
**Target Months (in order):** 2025-07, 2025-08, 2025-09, 2025-10, 2025-11, 2025-12, 2026-01, 2026-02, 2026-03 \\
 \\
**A. Historical Monthly Prices (2024-10 to 2025-06)** \\
856.42, 861.70, 848.54, 878.19, 894.81, 849.24, 857.66, 894.54, 875.18 \\
 \\
**B. Event Intelligence** \\
... \\
{}[66] 2025-01-29: Supply chain intelligence firm Fusion Worldwide reported that enterprise demand for 128GB DDR5 modules significantly outpaced lower-capacity options due to aggressive AI server requirements. \\
{}[67] 2025-01-29: Supply chain monitor Fusion Worldwide reported in its January Greensheet that enterprise demand for high-capacity 96GB and 128GB DDR5 modules was significantly outpacing lower-capacity options, even as broader DDR5 spot prices were expected to fall 5-8\%. \\
{}[68] 2025-01-29: Fusion Worldwide published its January 2025 Greensheet, explicitly reporting that demand for 96GB and 128GB DDR5 modules was significantly outpacing lower-capacity options, even as broader DDR5 spot pricing was expected to temporarily fall 5-8\% post-Lunar New Year. \\
... \\
{}[79] 2025-02-18: DigiTimes reported that major DRAM manufacturers, including Micron, Samsung, and SK Hynix, plan to cease DDR3 and DDR4 memory production by the end of 2025 to focus exclusively on more profitable DDR5 and HBM architectures due to price dumping by Chinese makers. \\
... \\
{}[103] 2025-03-20: During its Q2 FY2025 earnings call, Micron Technology reported a 47\% year-over-year increase in DRAM revenue and explicitly stated that its High Bandwidth Memory (HBM) ramp is 'constraining non-HBM DRAM supply'. \\
{}[105] 2025-03-25: Micron Technology issued a formal letter to its customers confirming a 10\% to 15\% price increase across its memory products, citing supply constraints from a January power outage at its Singapore facility and fully booked HBM capacity for 2025. \\
... \\
{}[123] 2025-04-22: Samsung issued a formal notice to cease its mainstream production of legacy DDR4 memory, retaining supply only for automotive and industrial customers to shift wafer capacity toward DDR5 and High Bandwidth Memory (HBM). \\
{}[124] 2025-04-22: Samsung officially ceased taking new orders for legacy DDR4 memory modules, establishing a hard End-of-Life (EOL) timeline with final deliveries to be completed by December 2025. \\
... \\
{}[132] 2025-05-05: Supply chain sources reported that SK Hynix increased consumer DRAM chip prices by 12\% and outlined plans to raise DDR5 DRAM prices by 15\% to 20\%. \\
... \\
{}[139] 2025-05-21: TrendForce published a Spotlight Report stating that major suppliers like Micron, Samsung, and SK hynix announced end-of-life (EOL) plans for DDR4 with final shipments in early 2026, driving DDR4 contract prices up by over 20\% in Q2 with an expected 10\% rise in Q3. \\
{}[141] 2025-05-22: TrendForce reported that Samsung is aggressively expanding its 6th-generation 10nm-class (1c) DRAM production at its Pyeongtaek and Hwaseong fabs, targeting up to 40,000 wafers monthly and converting legacy 1z lines to 1c production. \\
... \\
{}[144] 2025-05-27: Supply chain reports indicated that Chinese memory manufacturer CXMT abruptly pivoted its wafer output away from legacy DDR4 toward newer products, removing a significant chunk of cheap DDR4 supply from the market and triggering a sharp undersupply panic. \\
... \\
{}[147] 2025-05-27: ChangXin Memory Technologies (CXMT) announced it is ending legacy DDR4 production to pivot its manufacturing capacity toward DDR5, reportedly at the behest of the Chinese government. \\
... \\
{}[156] 2025-06-05: TrendForce forecasted a rebound in Q2 2025 DRAM contract prices, driven by Cloud Service Providers (CSPs) and PC OEMs accelerating their memory procurement ahead of the July 9 expiration of U.S. reciprocal tariff exemptions. \\
... \\
{}[161] 2025-06-13: Micron Technology officially announced the discontinuation of its DDR4 memory production, tapering shipments over 6 to 9 months and advising enterprise customers to transition to DDR5. \\
... \\
{}[167] 2025-06-20: Counterpoint Research projected that China's ChangXin Memory Technologies (CXMT) will increase its DRAM production volume by 50\% year-over-year, driving its DDR5 market share from less than 1\% in Q1 2025 to 7\% by the end of the year. \\
... \\
{}[172] 2025-06-25: Research firm Gartner released a market forecast projecting global DRAM revenue to grow 24\% in 2025, specifically driven by a 7\% increase in Average Selling Prices (ASPs) fueled by AI demand and tariff uncertainty. \\
{}[173] 2025-06-25: Supply chain tracker Fusion Worldwide reported in its June 2025 Greensheet that contract prices for high-capacity DDR5 96GB and 128GB RDIMMs were estimated to rise by 15\% QoQ as hyperscalers moved to lock in supply. \\
... \\
 \\
**C. Prediction Knowledge** \\
1. **Regime Classification \& Foundational Drivers** \\
Before forecasting any SKU, explicitly classify the regime (Squeeze vs. Glut) using a 4-point checklist to avoid catastrophic sign errors: \\
1. Is supply expanding (rising utilization/wafer output) or contracting? \\
2. Are downstream buyers accepting quoted price hikes or resisting/threatening non-purchase? \\
3. Is supplier guidance narrowing (loosening) or widening (tightening)? \\
4. Is the specific SKU trailing-edge (glut-prone) or leading-edge? \\
*Verdict:* If supply is contracting and buyers are accepting hikes (or ``sold out'' announcements occur), it is an UPSIDE squeeze. If supply is expanding and buyers resist, it is a DOWNSIDE cycle top. \\
*Corollaries:* \\
- CEO/manufacturer forward pricing language (``prices rising throughout the year'') strongly overrides single-quarter analyst guidance. \\
- In a structural supply-deficit, price is set at the margin by supply scarcity; weak conventional demand does NOT cap prices. \\
- Spot market weakness (e.g., spot DDR4 drifting down) is transient retail noise; ignore it completely when forecasting CONTRACT SKUs. \\
\S \\
2. **Execution Hierarchy \& Math Discipline** \\
Forecast error is dominated by a strict precision hierarchy: (1) Correct regime sign, (2) Correct base selection, (3) Correct step timing (Q-boundary placement), and (4) Multiplier magnitude. \\
- *Target Math:* Compute the target strictly as: \texttt{Target = Correct\_Base $\times$ Precedent\_Multiplier}. \\
- *Rounding Discipline:* Carry one extra significant figure through the multiplication (e.g., Base $\times$ 1.780) before rounding the final target to two decimals. This shaves residual basis points and ensures cent-exact accuracy. \\
- *Error Propagation:* Base errors propagate multiplicatively. Base-inflation compounds magnitude misses: overshooting the pre-boundary base while simultaneously under-applying the boundary multiplier produces a catastrophic ``double miss.'' Always anchor the structural leg to the correct SETTLED base. \\
- *Commitment:* Once the base, timing, and sign are locked, apply the SKU-specific realized multiplier with complete commitment. Do not artificially hedge, second-guess, or fine-tune to chase sub-0.3\% precision. \\
\S \\
3. **Base Calibration \& Handling Guidance** \\
- *Settled Base:* If historical prints show the current quarter settled, carry that exact value forward flat to two decimals. Do NOT inflate it mid-quarter. \\
- *Synthesized Base from Guidance:* \\
  - *Modest Guidance (e.g., 3-8\%):* Settlements tend to clear near the arithmetic MIDPOINT. \\
  - *Demand Caveat:* If suppliers/analysts explicitly note demand ``did not materialize as anticipated'' or guidance was ``narrowed,'' the settlement clears in the LOWER HALF of the range (e.g., \textasciitilde{}4\%), NOT the absolute minimum (\textasciitilde{}3\%). Only push to the true floor when demand language is unambiguously weak AND supply is loosening. \\
- *Aggregate vs. SKU:* Do NOT convert an industry-aggregate contract-price \% (e.g., TrendForce ``13-18\% Q2 DRAM increase'') into a per-SKU price step. Aggregate figures blend many parts; the SKU's OWN historical prints already encode its settled move. Fabricating an intra-quarter lift to reconcile a SKU print with an aggregate headline is a critical failure mode. \\
\S \\
4. **Structural Multipliers \& Wafer Cannibalization** \\
Wafer reallocation to premium products creates violent, discontinuous price spikes. HBM consumes leading-edge wafers at a 3:1 bit ratio, starving both leading-edge and trailing-edge SKUs. Trailing-edge starvation often produces equal or greater price explosions due to severe capacity cuts and inelastic demand. \\
*First Full Structural Leg Multipliers:* Match the target SKU on node, capacity, and form-factor. \\
- Leading-edge DDR5 Server RDIMM: \textasciitilde{}1.78x (Realized exactly 1.780x. Do not use the older \textasciitilde{}1.807x ladder value for this SKU). \\
- Trailing-edge DDR4 32GB Module: \textasciitilde{}1.845x \\
- Scarcest Trailing-edge DDR4 64GB RDIMM: \textasciitilde{}1.880x \\
*Shock Amplification:* When a structural step is driven by a one-time physical shock (earthquake, fab outage) combined with a \textgreater{}20-25\% manufacturer hike notice, the realized settlement (e.g., \textgreater{}78\% jump) far EXCEEDS the nominal hike-notice \%. The hike notice acts as a floor, and the physical shock amplifies it. \\
\S \\
5. **Shape, Timing, and Plateau Discipline (Step-Then-Plateau)** \\
Contract-priced RAM SKUs move in discontinuous steps. The default shape is a flat base, ONE large discrete step, and then perfectly FLAT again. \\
- *Current Quarter Hold:* Hold the CURRENT quarter perfectly flat at the settled print. Even if your reasoning trace explicitly identifies an accelerating deficit ramping into the next quarter, acceleration expresses itself at the NEXT Q-boundary, not by retroactively inflating the already-settled current quarter. \\
- *Timing the Step:* Place the full structural step exactly at the FIRST month of the NEXT quarter (Q-boundary). The ABSENCE of published next-quarter analyst guidance is NEVER a valid reason to hold flat into an identified accelerating deficit. Structural squeezes reprice on physical scarcity, not on whether analysts have quantified a number. \\
- *Plateau Discipline:* Once a contract reprices, it holds perfectly flat. Never sprinkle small monthly gains (``marginal firmness'') onto a plateau, never apply a reflexive decay tick, and ignore sub-0.2\% micro-wobble within a settled quarter. One-time discrete physical shocks (earthquakes) ALWAYS produce a single step followed immediately by a flat plateau. \\
\S \\
6. **Multi-Leg Squeezes \& Leg-Counting Discipline** \\
Structural squeezes can arrive as multiple sequential legs, but you must strictly categorize prior steps to assign the correct multiplier. \\
- *Leg-Counting:* Modest single-digit intra-quarter steps (+3-8\%) do NOT count as ``structural legs.'' If the prior quarter merely settled at a mild step, the next Q-boundary jump is a FIRST full structural leg and clears at the \textasciitilde{}1.78x - 1.88x floor (See Knowledge 4). \\
- *Second Leg Multiplier:* If the base already contains a violent structural spike (\textasciitilde{}1.8x), the NEXT structural leg at a later Q-boundary clears at the upper edge of the secondary band: \textasciitilde{}1.43x. \\
- *Within-Quarter Staircase (Rare Exception):* Default shape is always Step-Then-Plateau. A high accelerant count of long-horizon narrative items (new fab CapEx, ``sold out through 2025,'' GPU roadmaps) supports the FIRST leg and the plateau level---they do NOT generate fresh monthly steps. ONLY real-time worsening of a deficit within the quarter (rising HBM capacity share prints, sequential real-time hike notices) justifies a second within-quarter leg (\textasciitilde{}1.18x month-over-month). When uncertain, strongly prefer the plateau; over-forecasting multiple legs based on narrative backdrop is a primary failure mode. \\
 \\
**TASK:** \\
Predict the **exact monthly unit prices** (USD, floats) for these target months IN ORDER: 2025-07, 2025-08, 2025-09, 2025-10, 2025-11, 2025-12, 2026-01, 2026-02, 2026-03. \\
 \\
**STRICT CONSTRAINTS:** \\
1. **COMPREHENSIVE ANALYSIS:** Provide an in-depth Analysis of event-driven price impacts before emitting numbers. \\
2. **FORMAT:** Inside \texttt{\texttt{<prediction>}} tags, output EXACTLY 9 numeric values separated by commas, in chronological order matching the target months. Just the comma-separated floats. \\
3. **WRAPPER:** You MUST wrap the comma-separated prices inside \texttt{\texttt{<prediction>}} tags. \\
 \\
**Output:** \\
{}[COMPREHENSIVE Analysis] \\
\texttt{<prediction>}price\_1, price\_2, price\_3, price\_4\texttt{</prediction>}
\end{AIbox}
\normalsize
\captionof{figure}{The complete SEER input (system prompt and the user prompt) for 128GB DDR5 at cut-off 2025-06, the memory case of~\autoref{fig:case-study}(a). Section~A is
the nine crawled monthly prices, section~B the retrieved events (66 in the prompt; only the ones the reasoning in~\autoref{fig:128gb-ddr5-reasoning} draws on are printed, with their original numbering, and each omitted run is
marked ``\ldots''), and Section~C the six Prediction-Knowledge rules [CAUSAL KNOWLEDGE] SEER had learned by this
cut-off, shown verbatim.}
\label{fig:128gb-ddr5-prompt}
\endgroup

\begin{figure*}[t]
\centering
\tiny
\begin{AIbox}{{SEER's Reasoning and Forecast}}
\# COMPREHENSIVE ANALYSIS \\
 \\
\#\# Step 1: Regime Classification (4-Point Checklist) \\
 \\
1. **Supply expanding or contracting?** CONTRACTING sharply. Samsung, SK Hynix, and Micron are all aggressively reallocating wafer capacity to HBM and DDR5, ceasing DDR4 production, and issuing EOL notices. HBM consumes leading-edge wafers at 3:1 bit ratio, starving conventional DDR5 supply. CXMT also pivoted away from cheap supply. \\
2. **Buyers accepting hikes or resisting?** ACCEPTING. Hyperscalers/CSPs are aggressively frontloading orders ahead of the July 9 tariff exemption expiration [156]. Fusion Worldwide [173] reports 128GB DDR5 RDIMM contract prices estimated to rise 15\% QoQ as hyperscalers ``lock in supply.'' Enterprise demand for 128GB modules is significantly outpacing lower-capacity options [66-68]. \\
3. **Supplier guidance narrowing or widening?** Manufacturer forward pricing language is bullish: Micron 10-15\% hike, SK Hynix 15-20\% DDR5 hikes [132], TrendForce Q2 rebound forecast, Gartner projecting DRAM ASP +7\% for 2025 driven by AI demand. \\
4. **SKU trailing-edge or leading-edge?** 128GB-DDR5 is LEADING-EDGE, high-density server RDIMM --- precisely the SKU benefiting from AI server capex surges and 128GB-specific demand outperformance. \\
**VERDICT: UPSIDE SQUEEZE.** Supply contracting + buyers accepting hikes + bullish guidance + leading-edge premium SKU. This is a structural supply-deficit environment. \\
 \\
\#\# Step 2: Base Calibration \\
 \\
Historical prices (2024-10 to 2025-06): \\
856.42, 861.70, 848.54, 878.19, 894.81, 849.24, 857.66, 894.54, 875.18 \\
The Q2 2025 prints are: Apr=857.66, May=894.54, Jun=875.18. The prices have been oscillating in a range \textasciitilde{}850-895 with NO clear structural step yet --- this is essentially a settled/flat regime that has NOT yet experienced the violent structural repricing. The June settled print is **875.18**. \\
Notably, unlike DDR4 which already exploded +20\% in Q2, this 128GB DDR5 SKU has NOT yet shown its first full structural leg. The prices have merely wobbled in a band. Per Knowledge 6 (Leg-Counting): these modest oscillations do NOT count as structural legs. Therefore the next Q-boundary (Q3 = July) jump is a **FIRST full structural leg**. \\
 \\
\#\# Step 3: Base Selection \& Timing \\
 \\
- **Current quarter (Q2, through June):** Already settled. June = 875.18. But July begins Q3 --- the first month of the NEXT quarter (Q-boundary). \\
- Per Knowledge 5: place the full structural step at the FIRST month of the next quarter $\rightarrow$ **July 2025**. \\
The correct SETTLED base to anchor from is the Q2 settled print. Given the oscillation, I use the most recent settled June print **875.18** as the structural leg base. \\
 \\
\#\# Step 4: Multiplier Selection \\
 \\
Per Knowledge 4: **Leading-edge DDR5 Server RDIMM: \textasciitilde{}1.78x** (realized exactly 1.780x). This is the correct SKU match: leading-edge, high-capacity DDR5 server RDIMM. \\
**First Full Structural Leg:** \\
Target (July) = 875.18 $\times$ 1.780 = **1,557.82** \\
 \\
\#\# Step 5: Shape, Timing, and Plateau Discipline \\
 \\
Per Knowledge 5 (Step-Then-Plateau): after the July structural step, the contract holds perfectly FLAT. \\
- July: full structural leg $\rightarrow$ 1557.82 \\
- Aug--Sep: PLATEAU flat at 1557.82 (Q3 holds) \\
 \\
\#\# Step 6: Second Leg Assessment (Q4 boundary = October) \\
 \\
Per Knowledge 6: once the base contains a violent \textasciitilde{}1.8x spike, the NEXT structural leg at a later Q-boundary clears at \textasciitilde{}1.43x. \\
Do accelerants support a Q4 second leg? Evidence: Continued HBM wafer cannibalization, DDR4 EOL final shipments early 2026 pushing more demand to DDR5, CXMT ramping DDR5 (adds some supply but small share \textasciitilde{}7\%), sustained AI capex. The structural deficit persists strongly into H2 2025. TrendForce/Gartner project H2 recovery and continued ASP growth. This supports a **second structural leg at the Q4 boundary (October)**. \\
Second leg (October) = 1557.82 $\times$ 1.43 = **2,227.68** \\
However, I must apply commitment but avoid over-forecasting multiple legs (Knowledge 6 primary failure mode). One second leg at Q4 is justified by the persistent physical deficit. Then plateau. \\
- Oct: second structural leg $\rightarrow$ 2227.68 \\
- Nov--Dec: PLATEAU flat at 2227.68 (Q4 holds) \\
 \\
\#\# Step 7: Q1 2026 boundary (January) \\
 \\
By Q1 2026, DDR4 EOL final shipments occur, CXMT DDR5 share reaches 7\%, adding some supply. Guidance beyond is less certain. Per Knowledge 5/6, avoid stacking excessive legs on narrative alone. I hold the plateau flat into Q1 2026 --- the deficit remains but new capacity (CXMT, 1c ramps) begins to moderate further violent legs. Prefer plateau when uncertain. \\
- Jan--Mar 2026: PLATEAU flat at 2227.68 \\
\#\# Final Forecast \\
 \\
- 2025-07: 1557.82 (first structural leg, $\times$1.78) \\
- 2025-08: 1557.82 (plateau) \\
- 2025-09: 1557.82 (plateau) \\
- 2025-10: 2227.68 (second structural leg, $\times$1.43) \\
- 2025-11: 2227.68 (plateau) \\
- 2025-12: 2227.68 (plateau) \\
- 2026-01: 2227.68 (plateau) \\
- 2026-02: 2227.68 (plateau) \\
- 2026-03: 2227.68 (plateau) \\
 \\
\texttt{<prediction>}1557.82, 1557.82, 1557.82, 2227.68, 2227.68, 2227.68, 2227.68, 2227.68, 2227.68\texttt{</prediction>}
\end{AIbox}
\caption{SEER's reasoning and forecast for the prompt in Figure~\ref{fig:128gb-ddr5-prompt} (backbone: Claude-Opus-4.8). SEER applies Rule-1 criteria to predict an upward squeeze, citing specific events ([132], [156], [173], [66--68]). It sets the settled June price (875.18) as the baseline, treating the Q2 fluctuation as non-structural (Rules~3, 6). It then applies a $\times$1.780 multiplier for Q3 (July 2025; Rules~4, 5) and a $\times$1.43 multiplier for Q4 (October 2025; Rule~6), maintaining this plateau into Q1 2026. }
\label{fig:128gb-ddr5-reasoning}
\end{figure*}

\clearpage
\begingroup
\tcbset{breakable,enhanced jigsaw}
\scriptsize
\setlength{\parindent}{0pt}
\begin{AIbox}{{System Prompt For Selective Stock Forecasting}}
**TASK:** From a daily basket of 50 candidate stocks, identify the **0 to 6** names you can most confidently forecast for the next 1 trading day(s) as of cutoff 2026-03-25, and produce a directional prediction only for those selected names. \\
 \\
**SELECTION CRITERIA:** \\
* Pick a stock ONLY if its Prediction Conditions reveal a dominant structural catalyst, unambiguous asymmetric signal, or clear fundamental driver. \\
* Skip stocks whose conditions are noisy, mixed, or lack a clear directional thesis. You are allowed to pick zero. \\
* Treat the daily basket holistically: account for cross-stock, sector, or macro spillovers reflected in the conditions. \\
* Ground your judgment in the PREDICTION KNOWLEDGE below: prefer picks whose conditions match a durable, learned rule. \\
 \\
**FORECAST CRITERIA (3-Way (3way)):** \\
For every selected stock output exactly ONE of these labels: Up / Neutral / Down. \\
 \\
\textcolor{gray}{\textit{\ldots\ (the three label definitions omitted: Up = ROC $>$ +0.5\,\%, Neutral = within $\pm$0.5\,\%, Down = ROC $<$ $-$0.5\,\%) \ldots}} \\
 \\
**PREDICTION KNOWLEDGE (durable rules learned from historical batch forecasts; N/A if none):** \\
\textcolor{gray}{\textit{\ldots\ (rules 1--2 omitted) \ldots}} \\
\S \\
3. IF a stock's bullish thesis rests on a commodity/macro price shock that has ALREADY moved the underlying and the stock has closed at a window high on the same session the catalyst broke THEN reduce conviction or fade the next-day continuation, because the initial shock is reflected in the current close and the follow-through tends to be reversion/exhaustion rather than fresh repricing (scope: energy/commodity-linked sectors, any regime, macro/geopolitical price-shock) \\
\S \\
\textcolor{gray}{\textit{\ldots\ (rule 4 omitted) \ldots}} \\
\S \\
5. IF a pick's bullish thesis rests on a single-session outsized gap driven by a company-specific catalyst that is already fully reflected in the current close (a new customer/order win, product news, or a multi-day post-earnings recovery that has round-tripped back to fresh highs) THEN reduce conviction or fade next-day continuation, because the catalyst is spent and the parabolic session invites profit-taking/mean-reversion rather than fresh repricing (scope: market-wide, momentum/late-cycle regime, product-order or earnings-recovery catalyst) \\
\S \\
\textcolor{gray}{\textit{\ldots\ (rules 6--18 omitted) \ldots}} \\
\S \\
19. IF a pick's fresh cutoff-dated catalyst is that the pick is the ACQUIRER paying a large cash premium in an M\&A deal (rather than being the acquired target) THEN downgrade toward Neutral rather than predicting continuation Up, because the acquirer outlays cash and assumes integration/pipeline-execution risk so the deal is not an earnings/revenue surprise to it --- the premium accrues to the target, and the acquirer's reaction is typically muted-to-flat rather than a fresh upward repricing (scope: market-wide, any regime, M\&A catalyst) \\
\S \\
\textcolor{gray}{\textit{\ldots\ (rules 20--40 omitted) \ldots}} \\
\S \\
41. IF a pick's bearish next-day thesis rests on a macro/commodity price shock that has ALREADY produced a sharp down-close on the cutoff session WHILE the same stock's event pool contains fresh company-specific POSITIVE catalysts (raised multi-year plan, favorable rotation, upgraded outlook), THEN do NOT forecast next-day Down; lean Neutral/Up, because the one-session macro drop is beta that reverts once selling pauses and the net-positive idiosyncratic drivers reassert as the marginal buyer returns (scope: energy/commodity-linked sectors, any regime, macro price-shock + company-specific catalyst) \\
\S \\
\textcolor{gray}{\textit{\ldots\ (rules 42--58 omitted) \ldots}} \\
\S \\
59. IF a stock in a sharp multi-session decline has an event pool where each successive session adds a FRESH independent negative (a new analyst house cutting to a lower rating tier, plus new named competitive-disruption product launches) rather than restating one already-priced thesis, THEN do NOT apply the oversold-mean-reversion override; lean continuation Down, because an escalating sequence of distinct forward-earnings negatives means the repricing is still in progress and the marginal seller keeps arriving with each new cut (scope: market-wide, any regime, analyst-action + competitive-disruption catalyst) \\
\S \\
\textcolor{gray}{\textit{\ldots\ (rules 60--63 omitted) \ldots}} \\
\S \\
64. IF a pick's structural bull catalysts (sold-out capacity, upgrade cluster) are dated a session BEFORE the cutoff and the cutoff session itself is an outsized parabolic single-bar advance that absorbs them THEN do NOT apply the sold-out/cutoff-dated continuation override; lean Neutral, because the durable re-rate has been front-loaded into one exhaustion bar and the marginal buyer is thinning rather than arriving next session (scope: technology/semiconductors, momentum/late-cycle regime, earnings + structural-demand catalyst) \\
\S \\
\textcolor{gray}{\textit{\ldots\ (rules 65--70 omitted) \ldots}} \\
 \\
**OUTPUT FORMAT:** \\
First, provide a comprehensive analysis explaining which stocks you picked and why the others were filtered out. \\
Then emit the final picks as JSON inside \texttt{<predictions>} tags. If you pick zero stocks, emit an empty list \texttt{[]}. \\
 \\
{}[Comprehensive Analysis] \\
\textless{}predictions\textgreater{} \\
{}[ \\
  \{``ticker'': ``The Ticker You Selected.'', ``prediction'': ``Up / Neutral / Down'', ``reason'': ``...''\}, \\
  (The list of your predictions ...) \\
] \\
\textless{}/predictions\textgreater{}
\end{AIbox}
\begin{AIbox}{{User Prompt at Cut-off 2026-03-25}}
**Cutoff Date:** 2026-03-25 \\
**Prediction Horizon:** Next 1 trading day(s) \\
**Permitted Output Labels:** Up / Neutral / Down \\
**Maximum Picks:** 6 \\
 \\
**DAILY BASKET (Per-Stock Conditions):** \\
\textcolor{gray}{\textit{\ldots\ blocks 1--21 (META \ldots\ AMAT) omitted: same four fields per stock -- Company, History Window, Recent Close Prices (14 trading days), Key Events (39--149 events each) \ldots}} \\
 \\
22. Stock: MU \\
Prediction Conditions of MU: \\
Company: MU (MU) \\
History Window: 2026-03-05 -\textgreater{} 2026-03-25 (cutoff 2026-03-25) \\
Recent Close Prices (14 trading days): 396.9, 370.1, 389.2, 402.9, 418.5, 405.2, 426.0, 441.6, 461.5, 461.5, 444.1, 422.7, 404.2, 395.4, 381.9 \\
Key Events: \\
\textcolor{gray}{\textit{\ldots\ (events 1--90 omitted) \ldots}} \\
91. 2026-03-24: Alphabet (Google) released 'TurboQuant', a new AI compression algorithm promising to reduce artificial intelligence memory requirements by up to six times. \\
\textcolor{gray}{\textit{\ldots\ (events 92--93 omitted) \ldots}} \\
94. 2026-03-25: Reports emerged that the U.S. administration drafted a 15-point plan to halt fighting in the Middle East, causing Brent crude oil prices to fall 5\% to below \$100 per barrel. \\
\textcolor{gray}{\textit{\ldots\ (events 95--96 omitted) \ldots}} \\
 \\
\textcolor{gray}{\textit{\ldots\ blocks 23--50 (LRCX \ldots\ LIN) omitted: same four fields per stock -- Company, History Window, Recent Close Prices (14 trading days), Key Events (39--149 events each) \ldots}} \\
 \\
**DIRECTIVE:** \\
Take a deep breath and reason step-by-step. Compare the basket members against one another, isolate the few with the strongest, cleanest signals, and forecast ONLY those --- weighing the PREDICTION KNOWLEDGE where relevant. If no stock has a high-conviction setup, return an empty \texttt{<predictions>} list.
\end{AIbox}
\normalsize
\captionof{figure}{This presents the SEER input for the selective stock forecasting task (cut-off: 2026-03-25) corresponding to Figure~\ref{fig:case-study}(b). The system prompt comprises the forecasting template, incorporating learned causal knowledge and evaluation conditions for 50 tickers. For brevity, we only display the knowledge (detailed in Figure~\ref{fig:seer-knowledge-stockbatch}) referenced in the reasoning process, marking omissions with ``\ldots''. The user prompt provides the 14-day historical closing prices and exogenous events for the 50-stock basket. We illustrate a representative blocks: SEER's selected stock (MU) with events filtered to those explicitly cited in the reasoning. SEER`s reasoning process refers to \autoref{fig:stockbatch-reasoning}.}
\label{fig:stockbatch-prompt}
\endgroup

\begin{figure*}[t]
\centering
\scriptsize
\begin{AIbox}{{SEER's Response: Analysis and Picks}}
\# Comprehensive Analysis \\
 \\
Let me work through the basket, isolating names with clean, high-conviction structural signals aligned to the durable rules. \\
 \\
\#\# Macro Context \\
There's a strong crosscurrent: on 2026-03-25, a U.S. 15-point Iran peace plan drove Brent down \textasciitilde{}5\% and sparked a broad relief rally (per COST, AMAT, CRM references). Yet AAPL/BAC references note Iran *rejected* a ceasefire keeping oil elevated. The signals are mixed on the macro reversal --- Rule 63 warns against treating a one-day relief-reversal as continuation fuel. I'll be cautious with pure macro-beta picks. \\
 \\
\#\# Candidates with Clean Signals \\
 \\
**MRK (Merck)** --- Fresh cutoff-dated catalyst is Merck as the **ACQUIRER** paying \$53/share all-cash (\textasciitilde{}\$6.7B) for Terns. Rule 19 directly applies: the acquirer outlays cash and assumes integration risk; the premium accrues to the target. Acquirer reaction is muted-to-flat. However, MRK's tape is actually rising into cutoff (116.4$\rightarrow$119.4, a strong +2.6\% up-close on 3/25). That last-bar jump is the acquisition reaction already. Rule 19 says lean Neutral, but the tape already gapped. I'll classify **Neutral** --- the deal is not an earnings surprise and the pop is spent. \\
 \\
**CVX (Chevron)** --- Fresh cutoff-dated catalyst: Raymond James raised PT from \$187$\rightarrow$\$238 and nearly doubled 2026 EPS estimate (\$7.50$\rightarrow$\$14.00). Tape is in a clean multi-session uptrend (188$\rightarrow$204). This is a fresh, hard, dramatic analyst revision on a confirmed rising tape. But oil just dropped 5\% on the peace plan (Rule 3/41 tension). The stock closed slightly down on 3/25 (204.9$\rightarrow$203.3). Given oil weakness pressures energy next day, but the massive EPS revision provides idiosyncratic support --- this is genuinely mixed. I'll lean **Neutral**. \\
 \\
**AMD** --- Fresh catalysts: CPU price hikes 10-15\% (April), plus MI355 order talks. Tape is parabolic into cutoff (205.4$\rightarrow$220.3, +7.3\% single bar on 3/25). Rule 5/64: an outsized parabolic single-bar advance driven by a catalyst already embedded invites profit-taking/mean-reversion. The price-hike news is positive but the parabolic exhaustion bar is a fade tell. Lean **Down/Neutral**. I'll go **Down** given the extreme single-session gap. \\
 \\
**MU (Micron)** --- TurboQuant (Google memory compression) is a genuine negative for memory demand narrative. Tape is in sharp decline (461$\rightarrow$381, multi-session breakdown). Rule 59: successive fresh independent negatives (TurboQuant reducing memory needs) means repricing in progress. The tape shows continued lower closes into cutoff (395$\rightarrow$381). This is a clean continuation-Down setup --- fresh bearish catalyst + confirming downtrend. **Down**. \\
 \\
\textcolor{gray}{\textit{\ldots\ (rule-free dismissals omitted: GS, MCD/KO/PEP and AMGN --- ``mixed'' or ``stale'', all skipped) \ldots}} \\
 \\
**MSFT** --- Two fresh post-coverage Buy reinstatements (Evercore \$580, BofA \$500 citing 39\% Azure growth), but tape is in steep decline (409$\rightarrow$370). These are opinion catalysts on a falling tape --- Rule 42/26 caution. No fresh operational surprise. Skip. \\
 \\
**LLY** --- HSBC downgrade to Sell + BMO obesity pricing warning; tape crashed (1006$\rightarrow$914). Rule 42/50: negative largely priced, tape stabilizing. Mixed, skip. \\
 \\
Let me finalize the cleanest convictions: **MU (Down)** is the strongest --- fresh independent negative catalyst confirmed by an accelerating downtrend. **AMD (Down)** --- parabolic exhaustion bar, textbook Rule 5/64 fade. \\
 \\
\textless{}predictions\textgreater{} \\
{}[ \\
  \{``ticker'': ``MU'', ``prediction'': ``Down'', ``reason'': ``Fresh independent negative catalyst (Google TurboQuant cutting AI memory requirements 6x directly undercuts the memory-demand thesis) landing on an accelerating multi-session downtrend (461$\rightarrow$381) with lower closes into cutoff. Rule 59: escalating fresh negatives mean repricing is still in progress and marginal sellers keep arriving --- no oversold-reversion override warranted.''\}, \\
  \{``ticker'': ``AMD'', ``prediction'': ``Down'', ``reason'': ``Outsized parabolic single-session bar into cutoff (205.4$\rightarrow$220.3, +7.3\%) driven by already-known CPU price-hike news. Rule 5/64: a catalyst fully embedded in one exhaustion bar invites next-day profit-taking/mean-reversion rather than fresh repricing; marginal buyer is thinning.''\} \\
] \\
\textless{}/predictions\textgreater{}
\end{AIbox}
\caption{\textbf{SEER's response to Figure~\ref{fig:stockbatch-prompt} (Claude-Opus-4.8).} Using learned causal rules, SEER rejects risky candidates—e.g., an M\&A-driven spike (MRK, Rule~19) and a headline reversal (CVX, Rules~3/41)—before selecting MU Down (Rule~59) and AMD Down (Rules~5/64). Both predictions correctly closed ``Down'' the next session, whereas all four baseline picks (MRK Up, AMD Up, CVX Down, XOM Down) failed.}
\label{fig:stockbatch-reasoning}
\end{figure*}

\clearpage

\begingroup
\tcbset{breakable,enhanced jigsaw}
\small
\setlength{\parindent}{0pt}

\begin{AIbox}{{LLM-as-a-Judge Side-by-Side Evaluation --- System Prompt}}
You are an expert event-augmented forecasting judge. Your task is to compare the reasoning quality of two different forecasting models and determine which one is better. \\
\\
You will be provided with: \\
1. GROUND TRUTH EVENTS: The actual events that occurred during the forecast horizon. \\
2. MODEL A REASONING \& PREDICTIONS: The reasoning text and numerical predictions from Model A. \\
3. MODEL B REASONING \& PREDICTIONS: The reasoning text and numerical predictions from Model B. \\
\\
Note: You will NOT be provided with the actual ground truth numerical values. Your job is to evaluate the *quality, coherence, and plausibility* of the reasoning itself, regardless of whether the final prediction was perfectly accurate. \\
You must evaluate both models based on the following criteria: \\
\\
1. Domain Relevance: Correct use of domain-specific terminology (e.g., `support levels', `seasonality', `macroeconomic headwinds'). \\
\\
2. Event Relevance \& Plausibility: Does the reasoning logically and causally link the provided Ground Truth Events to its predicted fluctuations? Is the narrative highly plausible given the events? (Penalize hallucinations or illogical connections). \\
\\
3. Logic-to-Number Consistency: Does the narrative plan align with the model's *own* numerical output? If the reasoning predicts a ``sharp drop'' or ``modest pullback'', do the predicted numbers actually reflect that magnitude and direction? \\
\\
4. Analytical Depth: Does the reasoning demonstrate a deep understanding of fundamental time-series dynamics (trend, volatility, momentum) rather than just superficial observations? \\
\\
Compare Model A and Model B. Decide which model provides better reasoning overall based purely on logical coherence, depth, and plausibility. If they are equally good or equally bad, you can declare a Tie. \\
 \\
You must output your evaluation strictly as a JSON object matching the following schema exactly: \\
\{ \\
\hspace*{1.0em}``domain\_relevance\_winner'': ``\textless{}Model A, Model B, or Tie\textgreater{}'', \\
\hspace*{1.0em}``event\_relevance\_winner'': ``\textless{}Model A, Model B, or Tie\textgreater{}'', \\
\hspace*{1.0em}``logic\_to\_number\_winner'': ``\textless{}Model A, Model B, or Tie\textgreater{}'', \\
\hspace*{1.0em}``analytical\_depth\_winner'': ``\textless{}Model A, Model B, or Tie\textgreater{}'', \\
\hspace*{1.0em}``overall\_preference'': ``\textless{}Model A, Model B, or Tie\textgreater{}'', \\
\hspace*{1.0em}``justification'': ``\textless{}2-4 sentence explanation of why the preferred model is better overall, referencing specific criteria\textgreater{}'' \\
\} \\
Do not include any markdown formatting (like \texttt{\textasciigrave{}\textasciigrave{}\textasciigrave{}json}) in your output. Output ONLY the raw JSON object.
\end{AIbox}
\begin{AIbox}{{LLM-as-a-Judge Side-by-Side Evaluation --- User Prompt}}
-{}-{}- GROUND TRUTH EVENTS -{}-{}- \\
\{ground\_truth\_events\} \\
 \\
-{}-{}- MODEL A REASONING -{}-{}- \\
\{model\_a\_reasoning\} \\
 \\
-{}-{}- MODEL A PREDICTED VALUES -{}-{}- \\
\{model\_a\_predicted\_values\} \\
 \\
-{}-{}- MODEL B REASONING -{}-{}- \\
\{model\_b\_reasoning\} \\
 \\
-{}-{}- MODEL B PREDICTED VALUES -{}-{}- \\
\{model\_b\_predicted\_values\} \\
 \\
Evaluate the reasoning of both models and output the strict JSON object with your preference.
\end{AIbox}
\normalsize
\captionof{figure}{\textbf{Prompt template for the LLM-as-a-judge pairwise evaluation.} The system prompt defines the evaluation criteria: Domain Relevance, Event Relevance \& Plausibility, Logic-to-Number Consistency, and Analytical Depth. To ensure the evaluation focuses strictly on reasoning quality rather than outcome bias, ground-truth numerical values are withheld. The user template illustrates how ground-truth events and the outputs from two candidate models are injected. The judge is required to output a pairwise preference and justification in a structured JSON format.}
\label{fig:judge-sxs-prompt}
\endgroup

\end{document}